\documentclass[sigconf]{acmart}

\usepackage{subfigure}
\usepackage{bm}
\usepackage{color}
\usepackage{subfigure}
\usepackage{bm}
\usepackage{caption}
\usepackage{booktabs}
\usepackage{soul}
\usepackage{multirow}
\usepackage{algorithmic}
\usepackage{graphicx}
\usepackage{textcomp}
\usepackage{xcolor}

\newcommand{\model}{\textsc{CORAL}}
\newcommand{\nop}[1]{}

\AtBeginDocument{%
  \providecommand\BibTeX{{%
    \normalfont B\kern-0.5em{\scshape i\kern-0.25em b}\kern-0.8em\TeX}}}

\newcommand\blfootnote[1]{%
  \begingroup
  \renewcommand\thefootnote{}\footnote{#1}%
  \addtocounter{footnote}{-1}%
  \endgroup
}

\copyrightyear{2023}
\acmYear{2023}
\setcopyright{acmlicensed}\acmConference[KDD '23]{Proceedings of the 29th ACM SIGKDD Conference on Knowledge Discovery and Data Mining}{August 6--10, 2023}{Long Beach, CA, USA}
\acmBooktitle{Proceedings of the 29th ACM SIGKDD Conference on Knowledge Discovery and Data Mining (KDD '23), August 6--10, 2023, Long Beach, CA, USA}
\acmPrice{15.00}
\acmDOI{10.1145/3580305.3599392} \acmISBN{979-8-4007-0103-0/23/08}

\begin{document}



\title[Disentangled Causal Graph Learning for Online Unsupervised RCA]{Disentangled Causal Graph Learning for \\Online Unsupervised Root Cause Analysis}


\author{Dongjie Wang\textsuperscript{\rm 1,2,\dag}, Zhengzhang Chen\textsuperscript{\rm 1,\S}, 
  Yanjie Fu\textsuperscript{\rm 2}, 
  Yanchi Liu\textsuperscript{\rm 1}, and Haifeng Chen\textsuperscript{\rm 1}} 
    \affiliation{ 
    \textsuperscript{\rm 1}\institution{NEC Laboratories America Inc.}
      \country{}
         \textsuperscript{\rm 2}\institution{University of Central Florida}
         \country{}
    \textsuperscript{\rm 1}\{dwang, zchen, yanchi, haifeng\}@nec-labs.com\\
     \textsuperscript{\rm 2}yanjie.fu@ucf.edu
    }





\begin{abstract}
The task of root cause analysis (RCA) is to identify the root causes of system faults/failures by analyzing system monitoring data. 
Efficient RCA can greatly accelerate system failure recovery and mitigate system damages or financial losses. However, previous research has mostly focused on developing offline RCA algorithms, which often require manually initiating the RCA process, a significant amount of time and data to train a robust model, and then being retrained from scratch for a new system fault.  

In this paper, we propose \model, a novel online RCA framework that can automatically trigger the RCA process and incrementally update the RCA model. \model\ consists of Trigger Point Detection, Incremental Disentangled Causal Graph Learning, and Network Propagation-based Root Cause Localization. The Trigger Point Detection component aims to detect system state transitions automatically and in near-real-time. To achieve this, we develop an online trigger point detection approach based on multivariate singular spectrum analysis and cumulative sum statistics. To efficiently update the RCA model, we propose an incremental disentangled causal graph learning approach to decouple the state-invariant and state-dependent information. After that, \model\ applies a random walk with restarts to the updated causal graph to accurately identify root causes. The online RCA process terminates when the causal graph and the generated root cause list converge. Extensive experiments on three real-world datasets demonstrate the effectiveness and superiority of the proposed framework.
\blfootnote{$\dag$Work done during an internship at NEC Laboratories America.}  \blfootnote{$\S$Corresponding author.}

\end{abstract}

\maketitle
\vspace{-10pt}
\section{Introduction}
\label{sec:intro}
Root Cause Analysis (RCA) aims to identify the underlying causes of system faults (a.k.a., anomalies, malfunctions, errors, failures) based on system monitoring data~\cite{andersen2006root,kiciman2005root}. 
RCA has been widely used in IT operations, telecommunications, industrial process control, {\em etc.}, because a fault in these systems can greatly lower user experiences, and cause huge losses. For instance, in 2021, an intermittent outage of Amazon Web Services 
caused over $210$ millions in losses~\cite{amazon_attack}.

\nop{Due to the complex connections in these large-scale systems, they are vulnerable to attack events and it is difficult to determine the underlying causes of system failures for system recovery.
It may cause catastrophic consequences and high expenditures.
For instance, in 2021, an incident event in the cloud service system of Amazon impacted many apps in the eastern United States, such as  airline reservations, auto payments, and video streaming services~\cite{amazon_attack}. 
Consequently, RCA in complex systems is significant and socially impactful.}

\begin{figure}[hbpt]
    \vspace{-0.2cm}
    \centering
    \includegraphics[width=0.82\linewidth]{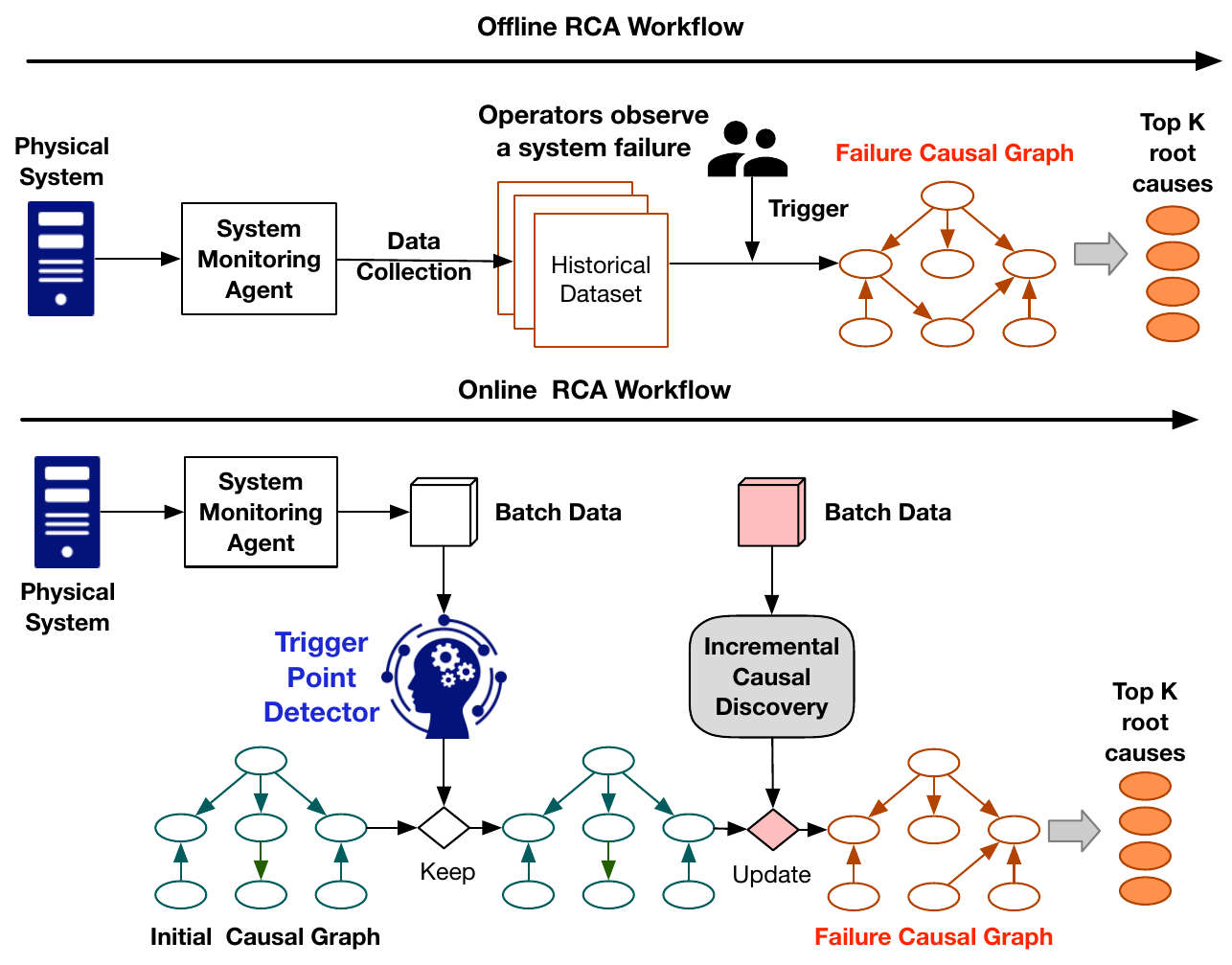}
    \captionsetup{justification=centering, singlelinecheck=off}
    \vspace{-0.15cm}
    \caption{Comparison between a traditional offline RCA workflow and the proposed \model\ online RCA workflow.
    }
    \vspace{-0.25cm}
    \label{fig:workflow}
\end{figure}

Previous RCA studies~\cite{liu2021microhecl,li2022causal,meng2020localizing,wang2023hierarchical} have focused primarily on developing effective \textbf{offline} methods for root cause localization. 
A key component of many data-driven offline RCA algorithms (see Fig.~\ref{fig:workflow}), especially the causal discovery based RCA methods~\cite{ikram2022root,li2022causal,meng2020localizing}, is to learn the causal structure or causal graph that profiles the causal relations between system entities and system Key Performance Indicator (KPI) based on historical data, so that the operators can trace back the root causes based on the built causal graph.
For instance, Ikram \textit{et al.}~\cite{ikram2022root} utilized historical multivariate monitoring data to construct causal graphs using the conditional interdependence test, and then applied causal intervention to identify the root causes of a microservice system.

However, there are three limitations in the traditional offline causal discovery based RCA workflow (\textbf{Fig.~\ref{fig:workflow}}). 
First, for a new system's faults, we need to retrain/rebuild the model from scratch. 
Second, the causal graph learning component is often time-consuming and requires a large amount of historical data to train a robust model.
Third, it often requires the operators to manually initiate the RCA process when they observe a system fault. As a result, it's often too late to mitigate any damage or loss caused by a system fault. 
These motivate us to ask the following questions: 
1) Is it possible to perform a causal discovery-based RCA task efficiently? 
2) How can we identify the root cause as early as possible? 
3) Can we deploy the RCA algorithm online for the streaming data? 
4) If deployed online, can we avoid the time-consuming retraining of the RCA model from scratch every time a system fault occurs? 

\nop{Therefore, a new perspective is necessary to perform the RCA problem with efficiency, fewer data requirements, and early detection.
Online RCA is an excellent answer to this question.
As shown in Figure~\ref{fig:workflow}, offline RCA has to manually collect historical data to construct the causal graph.
However, online RCA
can perform causal discovery on each new batch of monitoring data when finding early abnormal patterns.
This new formulation may detect the change in system states early with fewer data, resulting in more accurate and earlier diagnosis of underlying causes.}

In recent years, a very promising means for learning streaming data has emerged through the concept of incremental learning (\textit{aka} continual learning)~\cite{gepperth2016incremental}. Such incremental learning models rely on a compact representation of the already observed signals or an implicit data representation due to limited memory resources. In our RCA task, if we can incrementally update the RCA model or causal graph for each batch of the streaming data, it virtually accelerates the RCA process. More importantly, we don't need to wait until the system fault occurs to trigger the RCA process, or we might even be able to trigger an early RCA to mitigate the damages and losses. Thus, there exists a vital need for methods that can incrementally learn the RCA model and automatically trigger the RCA process. 

Enlightened by incremental learning, this paper aims to incrementally update the causal graph from streaming system monitoring data for accurately identifying root causes when a system failure or fault occurs. Formally, given the initial causal graph learned from historical data, and the streaming system monitoring data including entity metrics and KPI data, our goal is to automatically initiate the RCA process when a system fault occurs, incrementally update the initial causal graph by considering each batch of data sequentially, and efficiently identify the top $K$ nodes (\textit{i.e.}, system entities) in the updated causal graph that is most relevant to system KPI. There are two major challenges in this task:
\begin{itemize}
   \vspace{-0.1cm}
    \item \textbf{Challenge 1: Identify the transition points between system states to initiate root cause analysis.}
As aforementioned, in traditional RCA, the operators often manually initiate the root cause procedure after a system fault occurs. To mitigate the damages or losses, in an online setting, we need to automatically detect the system state changes caused by the system fault and trigger the RCA process. The challenge is how to identify the transition points of system states early if the fault does not affect the system KPI, but only affects some root cause system entities at the early stage. 

\item \textbf{Challenge 2: Incrementally update the causal graph model in an efficient manner.}
After the transition/trigger points are detected, we can not directly apply the old RCA model or  causal graph to identify the root causes, since the old causal graph only contains the causal relations learned from the previous system state data. Although some inherent system dependencies will never change over time~\cite{luo2018tinet} (\textit{i.e.}, system state-invariant causation), other causal dependencies may be highly dependent on the system state (\textit{i.e.}, system state-dependent causation). The challenge is how to identify the system state-invariant causation from the old model and quickly learn the state-dependent causation from the new batches of data for accelerating causal graph learning.

    \nop{As required by the online setting, the detecting method should be rapid with little delay.
    In the meantime, it may self-update its parameters automatically to fit the data of new states for more accurate detection results.}
    

    \nop{In real-world systems, the system state may shift over time due to system events (\textit{e.g.}, failure, shutdown), and the underlying causal graph may also shift accordingly.
    To capture such dynamics, we should rapidly learn the causal structures from monitoring data.
    \nop{However, existing causal discovery methods 
    are limited by: i) the time-consuming nature of causal learning; and
    ii) the need to retrain from scratch when  confronted with the data of new states.}}
    
\end{itemize}

To address these challenges, in this paper, we propose \model, a novel in\underline{c}remental causal graph learning framework, for \underline{o}nline \underline{r}oot c\underline{a}use \underline{l}ocalization. \model\ consists of  three main steps: 1) Trigger Point Detection; 2) Incremental Disentangled Causal Graph Learning; and 3) Network Propagation-Based Root Cause Localization. In particular, 
the first step of \model\ is to detect the transition points between system states in real time based on system entity metrics and KPI data.
To detect trigger points with less delay, we develop an online trigger point detection algorithm based on multivariate singular spectrum analysis and cumulative sum statistics.
These points are then used to trigger incremental causal graph learning. Assuming that as the system state transitions, the underlying causal structure partially changes and evolves over time instead of shifting abruptly and significantly.
Based on this assumption, we propose an incremental disentangled causal graph learning model to efficiently learn causal relations by decoupling state-invariant and state-dependent causations. 
After that, we apply a random walk with restarts to model the network propagation of system faults to accurately identify root causes. The online root cause localization process terminates for the current system fault when the learned causal graph and the generated root cause list converge. To summarize, our main contributions are three-fold:

\nop{a novel and practical online RCA framework, namely \model, by iterating incremental causal discovery and trigger point detection.
Specifically, there are three main steps:
1) trigger point detection; 2) incremental disentangled causal graph learning; 3) network propagation-based root cause localization.
The first core task of \model\ is to automatically detect the system state transition points.
To rapidly detect trigger points with less delay, we develop an online trigger point detection framework using multivariate singular spectrum analysis and cumulative sum statistics (CUSUM).
Then, these points are used to trigger the increment causal discovery.
Assuming that as the system state transitions, the underlying causal structure partially changes and evolves over time instead of shifting abruptly and significantly.
Thus, in step 2, based on this assumption, we propose an incremental disentangled causal graph learning model to efficiently learn causal relations by decoupling state-invariant and state-dependent causation. 
After that, in step 3, we apply a random walk with restarts to the well-built causal structure to model 
the network propagation of system faults to accurately identify root causes.
The online root cause localization process terminates for the current system event when the learned causal graph and the generated root cause list converge. Extensive experiments and case studies are conducted on three real-world datasets to validate the efficacy of our work.}

\begin{itemize}
   \vspace{-0.1cm}
    \item \textit{\textbf{Problem}}:
    We investigate the novel problem of online root cause localization. Remarkably, we propose to solve this problem by automatic trigger point detection and incremental causal structure learning.
    \nop{We propose a novel and practical online framework for root cause localization in order to perform system diagnosis using an online learning system with automated trigger point detection and incremental causal discovery.}
    \item \textit{\textbf{Algorithms}}: We propose a principled framework \model, which integrates a new family of disentangled representation learning (\textit{i.e.}, causal graph disentanglement), online trigger point detection, and incremental causal discovery.
    \nop{First, we propose a disentangled incremental causal learning model that decouples the state-invariant and state-dependent causation to efficiently learn causal relations over state transitions. 
    Second, we develop an online trigger point detection module using multivariate singular spectrum analysis in order to rapidly and accurately detect trigger points based on monitoring metric data.}
    
    \item \textit{\textbf{Evaluations}}:  We perform extensive experiments on three real-world datasets to validate the effectiveness of our approach. The experimental results demonstrate the superior performance of \model\ over the state-of-the-art methods on root cause localization.
\end{itemize}

\vspace{-0.2cm}
\section{Preliminaries }

\noindent \textbf{System Key Performance Indicator (KPI)} is a monitoring time series that indicates the system status. For example, in a microservice system, latency is a KPI to measure the system status. The lower (higher) a system's latency is, the better (worse) its performance is.

\noindent \textbf{Entity Metrics} are multivariate time series collected by monitoring numerous system entities/components. For example, in a microservice system, a system entity can be a physical machine, container, virtual machine, pod, and so on. 
The system metrics include CPU utilization, memory consumption, disk IO utilization, etc. 
System entities with anomalous metrics can be the root causes of abnormal system latency/connection time, which is a sign of a system fault.

\begin{figure*}[!t]
    \centering
    \includegraphics[width=0.95\linewidth]{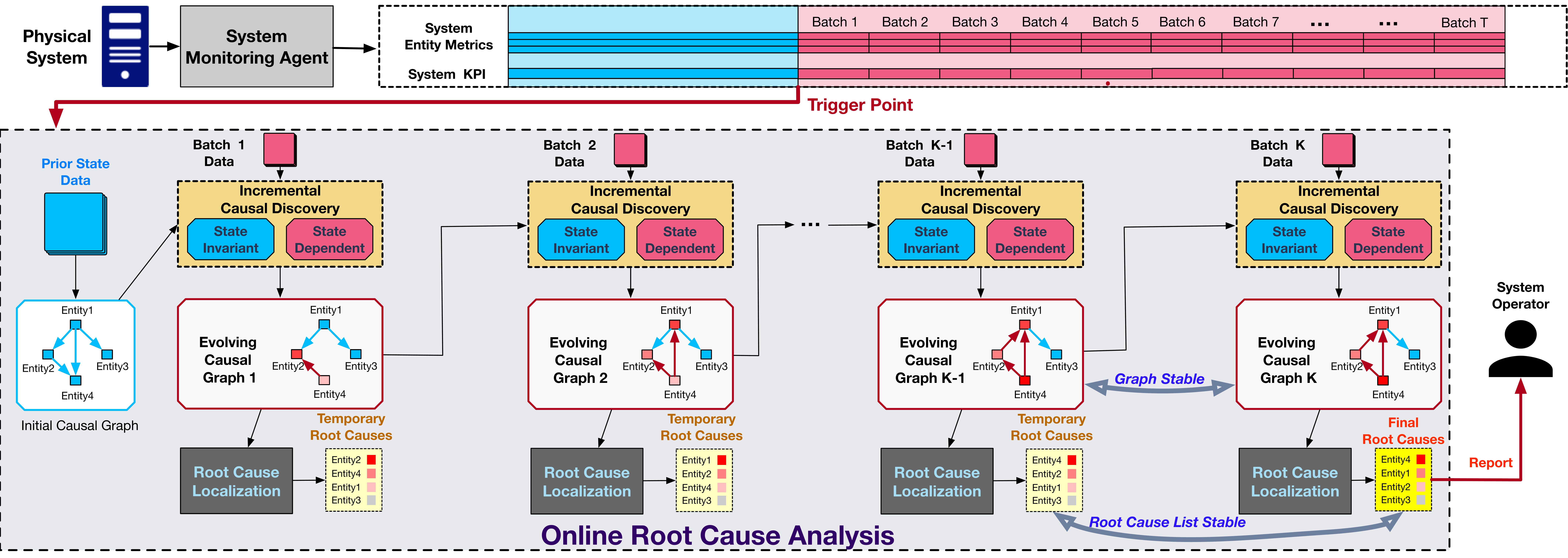}
    \captionsetup{justification=centering, singlelinecheck=off}
    \vspace{-0.2cm}
    \caption{The overview of the proposed framework \model. \model\ first detects trigger points using the monitoring entity metrics and KPI data.
    If detected, it will initiate incremental causal graph learning. Each data batch is used to incrementally update the previous causal graph by disentangling the state-invariant and state-dependent causations for RCA. When the learned causal graph and the root cause list converge, system operators will receive the final root causes for system recovery.
    }
    \vspace{-0.35cm}
    \label{fig:framework}
\end{figure*}

\noindent \textbf{Trigger Point or System State Change Point} is the time when the system transitions from one state to another. Real-world systems are dynamic. A system fault can cause a change in the system's status. As the state of a system varies, the underlying causal relationships between its components also change. Thus, to effectively identify root causes in an online setting, it is essential to learn different causal graphs in different states. From this perspective, the system state change points can be viewed as the triggers for updating the causal graph/online RCA model. 

\nop{Assuming that the underlying causal graph may evolve over time when the system state transits.
To identify root causes before causing serious results, 
we expect to detect the change points of system states.
These points will be used to trigger updating the causal graph incrementally for online RCA.
Our framework supports two
kinds of trigger points: 
1) the change points in the monitoring entity metrics and system KPI that are automatically detected by algorithms.
2) system operators can manually set the trigger point if the monitoring time series reveals early patterns of system events.}

\noindent \textbf{Problem Statement.}
Let $\mathcal{X} = \{ \mathbf{X}_1, \cdots, \mathbf{X}_N \}$ denote $N$ multivariate metric data.
The $i$-th metric data is $\mathbf{X}_i=[\mathbf{x}_1^i,\cdots,\mathbf{x}_T^i]$,
where $\mathbf{x}^i_t \in \mathbb{R}^{M}$ is the observation of $M$ system entities at time point $t$.
To reduce notational clutter, we omit the metric index $i$ and use $\mathbf{X}$ to represent $\mathbf{X}_i$ in the following sections. These observations are non-stationary, and the relationships between various system entities are dynamic and subject to change over time. We assume that the underlying state of the system can change when a system fault occurs, and the relationships between system entities in each state are represented by a directed acyclic graph (DAG).
For simplicity, we illustrate here using the system state transition from $s_p$ to $s_{p+1}$. The system KPI is $\mathbf{y}$. The monitoring entity metric data of $s_p$ is $\mathbf{\tilde{X}}_p \in \mathbb{R}^{\rho\times M}$, where $\rho$ is the time length in the state $s_p$. $G_p$ represents the causal graph of $s_p$, which consists of nodes representing system KPI or entities, and edges representing causal relations.
The data of state $s_{p+1}$ comes one batch at a time, denoted by $\mathbf{\tilde{X}}_{p+1}=[\mathbf{\check{X}}^1_{p+1},\cdots,\mathbf{\check{X}}^L_{p+1}]$, where the $l$-th batch $\mathbf{\check{X}}^l_{p+1} \in \mathbb{R}^{b \times M}$ and $b$ is the length of each batch of data.
Our goal is to automatically trigger the RCA process when a system fault occurs, incrementally update $G_p$ to $G_{p+1}$ by considering each batch of data sequentially, and efficiently identify the top $K$ nodes 
in the causal graph $G_{p+1}$ that are most relevant to $\mathbf{y}$. 

\vspace{-0.2cm}
\section{Methodology}
\textbf{Fig.~\ref{fig:framework}} shows the  framework \model\ for online root cause localization. Specifically, \model\ addresses the discussed challenges with three modules: (1) Online trigger point detection, which detects the transition points between system states and triggers incremental causal structure learning; (2) Incremental disentangled causal graph learning, a novel disentangled graph learning method that enables efficient causal structure learning by decoupling state-invariant and state-dependent causations; (3) Network propagation-based root cause localization, which models the network propagation of system faults to accurately identify the most possible root causes.

\begin{figure*}[!t]
    \centering
    \includegraphics[width=0.98\linewidth]{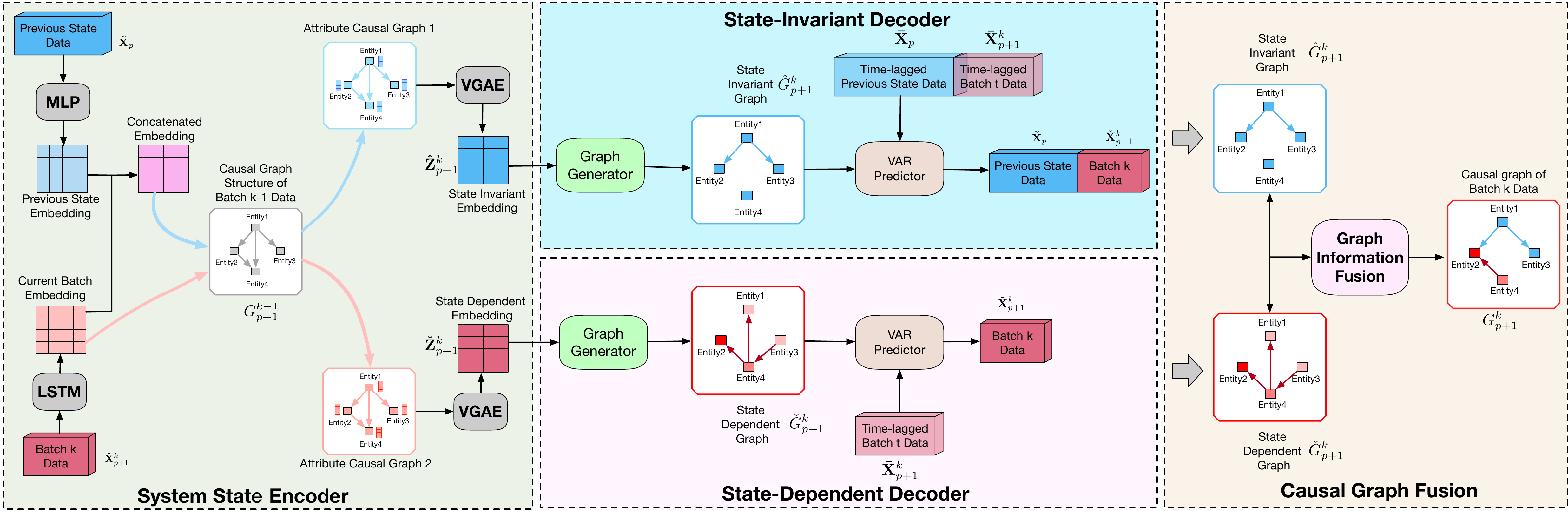}
    \captionsetup{justification=centering, singlelinecheck=off}
    \vspace{-0.2cm}
    \caption{Incremental Disentangled Causal Graph Learning. 
    The system state encoder aims to decouple state-invariant and state-dependent information in order to produce corresponding embeddings. State-invariant and state-dependent decoders aim to reconstruct and refine state-invariant and state-dependent causal relations, respectively.
    The causal graph fusion module aims to fuse state-invariant and state-dependent causation in order to obtain the new causal graph.
    }
    \vspace{-0.3cm}
    \label{fig:disentangle}
\end{figure*}


\subsection{Trigger Point Detection}
\label{tpd_sec}
\nop{System KPIs indicate a system's status or performance. An intuitive way is to apply a change point detection algorithm to the KPI data. However, the KPI data may not fully encode the system state changes caused by the system incident, especially in the RCA task when the root cause entity gradually causes changes in other entities with some propagation patterns or early abnormal symptoms. These early patterns of system state change might not be represented by the KPI data, but rather by entity metric data. 
Thus, our framework combines entity metric data and system KPI for trigger point detection. 
}

\nop{To start the increment causal discovery for online RCA, we have to detect the change points of system status.
There are two ways to find trigger points: 1) manually set up by system operators; 2) automatically detect by algorithms. 
Here, we will focus on the second method.
To satisfy the low-latency requirements of the online RCA scenario, we 
should provide an efficient detection algorithm.}

Our goal is to detect the trigger points by integrating both entity metrics and system KPI data. Inspired by~\cite{alanqary2021change}, we model the underlying dynamics of system entity metrics and KPI data (\textit{i.e.}, multivariate time series observations) through the Multivariate Singular Spectrum Analysis (MSSA) model. For simplicity, we add the system KPI, $\mathbf{y}$, as one dimension to the metric data, $\mathbf{X}$, to illustrate our model.

Specifically, given monitoring metric data $\mathbf{X}$, we first construct the base matrix, denoted by $\mathbf{Z}_{\mathbf{X}}$, by using the previous $T_0$ records.
The requirement for the initial value of $T_0$ here is that no system state transition occurs in the time segment $t \leq T_0$. 
The singular vectors of 
$\mathbf{Z}_{\mathbf{X}}$ are then grouped into two matrices $\mathbf{\hat{U}}_0$ and $\mathbf{\hat{U}}_\perp$.
We estimate the pre-change subspace $\mathbb{\hat{L}}_0$ as follows: 
\begin{equation}
    \mathbb{\hat{L}}_0 = \text{span}(\mathbf{\hat{U}}_0)
\end{equation}
Meanwhile, let $\mathbb{\hat{L}}_\perp=\text{span}(\mathbf{\hat{U}}_\perp)$ be the orthogonal complement of the subspace $\mathbb{\hat{L}}_0$.
After that, for the new data $t>T_0$, we build the L-lagged matrix $\mathbf{X}(t-L+1:t)$.
We compute the Euclidean distance between the L-lagged matrix and the estimated subspace $\mathbb{\hat{L}}_0$ as the detection score, which can be defined as: 
\begin{equation}
    D(t) = \left \| \mathbf{\hat{U}}_\perp^\top\mathbf{X}(t-L+1:t)  \right \|^2_F -c
\end{equation}
where $c\ge 0$ is the shift-downwards constant.
Moreover, we compute the cumulative detection score using the cumulative sum (CUSUM) statistics, which can be defined as:
\begin{equation}
    y(t) = \text{max}\{y(t-1)+D(t),0\} \quad y(T_0)=0
\end{equation}
if the $y(t)=0$, we proceed to check the next time point.
Otherwise, the change point is identified when $y(t)>h$, where $h$ is a predefined threshold. This process can be defined as:
\begin{equation}
    \hat{\tau} = \text{inf}_{t>T_0} \{t|y(t)\ge h\}
\end{equation}
The change point $\hat{\tau}$ is the trigger for incremental causal graph learning.
To recheck the next change point, the base matrix is updated with the time segment $\mathbf{X}(\hat{\tau},\hat{\tau}+T_0-1)$. This model can detect trigger points in nearly real-time.

\vspace{-0.1cm}
\subsection{ Incremental Disentangled Causal Graph Learning}
\nop{
 After a trigger point is detected, we propose a disentangled graph learning model to integrate system state invariant and system state-dependent information for incrementally constructing causal graphs. 
First, the system state has a strong influence on the relationships between system entities and their temporal dynamics. 
Second, some inherent system dependencies will never change over time. Thus, if we can learn state-invariant and state-dependent causal relations, we can accelerate the causal graph learning process.

We design an incremental  disentangled causal graph learning model, in which the system state encoder is to disentangle modeling the state-invariant and state-dependent embeddings;
state-invariant and state-dependent decoders are used to construct and refine the corresponding causal relations respectively;
causal graph fusion is to fuse the state-invariant and state-dependent causation to generate a new causal graph.
}

After a trigger point is detected,  we propose a disentangled causal graph learning model by integrating state invariant and state-dependent information, as well as incremental learning. 

\nop{Causal graph-based root cause analysis (RCA) methods have been proven effective in localizing root causes by capturing the propagation patterns of system faults~\cite{liu2021microhecl,wu2020microrca,li2022causal}.
There are two ways to collect the causal graph employed in these works: 1) created based on domain knowledge; 2) offline data-driven learning.
However, the following two limitations make them cannot be employed in the online setting:
1) the causal graph construction process is time-consuming and labor-intensive;
2) the causal learning process cannot capture complicated dynamics  of causal structures over system state transitions, leading to an inaccurate causal structure. 
Thus, we intend to disentangle the state-invariant and state-dependent causal relations among system entities  to quickly identify effective causal graphs.
To simplify the description, we adopt the system state transition event from $s_p$ and $s_{p+1}$ to show the disentangled causal graph learning process.}

\noindent\textbf{System State Encoder}.
The goal of the system state encoder is to aggregate the information from system state data and the corresponding causal graph.
We use the $k$-th batch data to illustrate our design. 
For simplicity, we assume that the previous state data $\mathbf{\tilde{X}}_{p}$ and the batch of new state data $\mathbf{\check{X}}_{p+1}^k$ both have the system KPI $\mathbf{y}$ as one dimension.

Given $\mathbf{\tilde{X}}_{p}$, $\mathbf{\check{X}}_{p+1}^k$ and the causal graph at the $k-1$ batch $G_{p+1}^{k-1}$, we aim to integrate their information into state-invariant and state-dependent embeddings, respectively.
First, we employ a fully-connected linear layer to preserve the information of $\mathbf{\tilde{X}}_{p}$ into a latent representation $\mathbf{U}_p$, which can be defined as: 
 \begin{equation}
      \mathbf{U}_p = \mathbf{\tilde{X}}_{p}\cdot\mathbf{W}_p + \mathbf{b}_p
 \end{equation}
where $\mathbf{W}_p$ and $\mathbf{b}_p$ are the weight matrix and bias item of the linear layer, respectively.

Then, to track the information change in the state $p+1$'s data batch, we employ a recurrent function $f(\cdot, \cdot)$. 
The function $f(\cdot , \cdot)$ takes $\mathbf{\check{X}}_{p+1}^k$ and the previous hidden state $\mathbf{H}^{k-1}_{p+1}$ as inputs and outputs a latent representation $\mathbf{H}^{k}_{p+1}$, which is defined as: 
\begin{equation}
    \mathbf{H}_{p+1}^{k} = f(\mathbf{\check{X}}_{p+1}^k,\mathbf{H}^{k-1}_{p+1})
\end{equation}
We use a long short-term memory network (LSTM)~\cite{hochreiter1997long} to implement $f(\cdot , \cdot)$.
Due to the computational overhead, we do not use the passed batch of data to update the new causal graph.
Thus, it is effective to track the causal dynamics using the LSTM module.

In addition, because state-invariant causal relations would be affected by both the previous state data and the new batch of data, whereas state-dependent causal relations are only affected by the new batch of data, to obtain the state-invariant embedding $\mathbf{\hat{Z}}_{p+1}^k$, we first concatenate $\mathbf{U}_p$ and $\mathbf{H}_{p+1}^k$ together, and then map it to the causal graph $G_{p+1}^{k-1}$ as the node embeddings.
After that, we employ the mapping function $g(\cdot , \cdot)$ to convert the attributed graph into the state-invariant embedding, which is defined as: 
\begin{equation}
    \mathbf{\hat{Z}}_{p+1}^k = g(\mathbf{A}_{p+1}^{k-1}, \text{Concat}(\mathbf{U}_p ,\mathbf{H}_{p+1}^k))
\end{equation}
where $\mathbf{A}_{p+1}^{k-1}$ represents the adjacency matrix of $G_{p+1}^{k-1}$ and Concat represents the concatenated operation.
to implement the function $g(.)$, we employ a variational graph autoencoder (VGAE)~\cite{kipf2016variational}. 
VGAE embeds all information into an embedding space that is smooth and continuous. This is helpful for capturing the causal dynamics between different system entities. To obtain the state-dependent embedding $\mathbf{\check{Z}}_{p+1}^{k}$, we only map $\mathbf{H}_{p+1}^k$ to $G_{p+1}^{k-1}$ as its attributes, and then employ another VGAE layer to convert the attributed graph to  the state-dependent embedding. 
This process can be defined as: 
\begin{equation}
    \mathbf{\check{Z}}_{p+1}^k = g(\mathbf{A}_{p+1}^{k-1}, \mathbf{H}_{p+1}^k)
\end{equation}

\noindent\textbf{State-Invariant Decoder.}
The goal of the state-invariant decoder is to learn the invariant causal relations across two system states. To recover the state-invariant part, we first feed $\mathbf{\hat{Z}}_{p+1}^k$ into the graph generator layer to generate the corresponding state-invariant graph $\hat{G}_{p+1}^{k}$, which is defined as: 
\begin{equation}
    \hat{G}_{p+1}^{k} = \text{Sigmoid}(\mathbf{\hat{Z}}_{p+1}^k \cdot \mathbf{\hat{Z}}_{p+1}^{k^\top})
\end{equation}
where Sigmoid is an activation function and $(.)^\top$ is a transpose operation. 
But since this process constructs the graph using the state-invariant embeddings only, it can't guarantee that the state-invariant causal relationships are shown accurately in this graph.
To overcome this issue, 
two optimization objectives must be met:
1) Making $\hat{G}_{p+1}^{k}$ as similar to the previous causal graph $G_{p+1}^{k-1}$ as possible.
2) Fitting $\hat{G}_{p+1}^{k}$ to both the previous and new state data batches.
To achieve the first objective, we minimize the reconstruction loss $\mathcal{L}_{\hat{G}}$, which is defined as: 
\begin{equation}
    \mathcal{L}_{\hat{G}} = \left \|\hat{\mathbf{A}}_{p+1}^{k}-\mathbf{A}_{p+1}^{k-1}  \right \|^2
\end{equation}
where $\hat{\mathbf{A}}_{p+1}^{k}$ and $\mathbf{A}_{p+1}^{k-1}$ are the adjacency matrices of $\hat{G}_{p+1}^{k}$ and $G_{p+1}^{k-1}$, respectively.
To achieve the second objective, we fit the graph and data with a structural vector autoregressive (SVAR) model~\cite{pamfil2020dynotears}. More specifically, given the time-lagged data of the previous state $\mathbf{\bar{X}}_p$ and the current new batch $\mathbf{\bar{X}}^k_{p+1}$, the SVAR-based predictive equations can be defined as:
\begin{equation}
\left\{
             \begin{array}{lr}
             \mathbf{\tilde{X}}_p=\mathbf{\tilde{X}}_p\cdot\mathbf{\hat{A}}_{p+1}^k + \mathbf{\bar{X}}_p\cdot\mathbf{\hat{D}}_{p+1}^k + \tilde{\epsilon}_p &  \\
             \mathbf{\check{X}}_{p+1}^k=\mathbf{\check{X}}_{p+1}^k\cdot\mathbf{\hat{A}}_{p+1}^k + \mathbf{\bar{X}}_{p+1}^k\cdot\mathbf{\hat{D}}_{p+1}^k + \check{\epsilon}_{p+1}^k
             \end{array}
\right.
\end{equation}
where $\tilde{\epsilon}_p$ and $\check{\epsilon}_{p+1}^k$ are vectors of centered error variables;
$\mathbf{\hat{A}}_{p+1}^k$ is used to capture causal relations among system entities;
and the weight matrix $\mathbf{\hat{D}}_{p+1}^k$ is used to model the contribution of time-lagged data for the predictive task.
$\mathbf{\hat{A}}_{p+1}^k$ and $\mathbf{\hat{D}}_{p+1}^k$ are used to predict both the past state data and the current batch of data.
To ensure the accuracy of learned causal structures, we minimize two predictive errors $\mathcal{L}_{\tilde{p}}$ and $\mathcal{L}_{\hat{p}}$, which are defined as:
\begin{equation}
\left\{
             \begin{array}{lr}
             \mathcal{L}_{\tilde{p}} = \left \| \mathbf{\tilde{X}}_p - (\mathbf{\tilde{X}}_p\cdot\mathbf{\hat{A}}_{p+1}^k + \mathbf{\bar{X}}_p\cdot\mathbf{\hat{D}}_{p+1}^k)  \right \|^2 &  \\
             \mathcal{L}_{\hat{p}} = \left \| \mathbf{\check{X}}_{p+1}^k - (\mathbf{\check{X}}_{p+1}^k\cdot\mathbf{\hat{A}}_{p+1}^k + \mathbf{\bar{X}}_{p+1}^k\cdot\mathbf{\hat{D}}_{p+1}^k) \right \|^2
             \end{array}
\right.
\end{equation}

\noindent\textbf{State-Dependent Decoder}
The goal of the state-dependent decoder is to learn the new causal relations introduced by the new batch of data.
Similar to the learning process of the state-invariant decoder, we first generate the state-dependent graph $\check{G}_{p+1}^{k}$ by applying the same strategy on the embedding $\mathbf{\check{Z}}_{p+1}^k$, which is defined as: 
\begin{equation}
    \check{G}_{p+1}^{k} = \text{Sigmoid}(\mathbf{\check{Z}}_{p+1}^k \cdot \mathbf{\check{Z}}_{p+1}^{k^\top})
\end{equation}
To ensure the causal graph $\check{G}_{p+1}^{k}$ is brought by the new batch of data $\mathbf{\check{X}}^{k}_{p+1}$, 
two optimization objectives must be met:
1) Making $\check{G}_{p+1}^{k}$  as similar to the complement of the previous causal graph as possible;
2) Fitting $\check{G}_{p+1}^{k}$ to the new batch of data. 
To achieve the first objective, we minimize the reconstruction loss $\mathcal{L}_{\check{G}}$, which is defined as: 
\begin{equation}
    \mathcal{L}_{\check{G}} = \left \|\check{\mathbf{A}}_{p+1}^{k}-(\sim\mathbf{A}_{p+1}^{k-1})  \right \|^2
\end{equation}
where $(\sim\mathbf{A}_{p+1}^{k-1})$ refers to the inversion of each element in the adjacency matrix $\mathbf{A}_{p+1}^{k-1}$ and $\check{\mathbf{A}}_{p+1}^{k}$ is the adjacency matrix of  $\check{G}_{p+1}^{k}$.
To achieve the second objective, we define a predictive equation using SVAR: 
\begin{equation}
\mathbf{\check{Y}}_{p+1}^k=\mathbf{\check{X}}_{p+1}^k\cdot\mathbf{\check{A}}_{p+1}^k + \mathbf{\bar{X}}_{p+1}^k\cdot\mathbf{\check{D}}_{p+1}^k + \epsilon
\end{equation}
where $\mathbf{\check{A}}_{p+1}^k$ captures the new causal relations introduced by the new data batch $\mathbf{\check{X}}_{p+1}^k$; and $\mathbf{\check{D}}_{p+1}^k$ is to model the contribution of time-lagged data for prediction. To ensure the accuracy of learned causal structures, we minimize the predictive error $\mathcal{L}_{\check{p}}$, which is defined as:
\begin{equation}
     \mathcal{L}_{\check{p}} = \left \| \mathbf{\check{X}}_{p+1}^k - (\mathbf{\check{X}}_{p+1}^k\cdot\mathbf{\check{A}}_{p+1}^k + \mathbf{\bar{X}}_{p+1}^k\cdot\mathbf{\check{D}}_{p+1}^k) \right \|^2
\end{equation}

\noindent\textbf{Causal Graph Fusion.}
From the state-invariant decoder and state-dependent decoder, we can obtain the state-invariant causal graph $\hat{G}_{p+1}^{k}$ and the state-dependent causal graph $\check{G}_{p+1}^{k}$, respectively.
To generate the causal graph $G_{p+1}^{k}$ for the current batch of data, simple addition will not work because it may result in dense and cyclical graphs. 
Here, we propose a new graph fusion layer to fuse the two causal graphs, which can be formulated as follows:
\begin{equation}
    \mathbf{A}_{p+1}^{k} = \text{RELU}(\text{tanh}(\mathbf{\hat{A}}_{p+1}^{k}\cdot\mathbf{\check{A}}_{p+1}^{k^\top} - \mathbf{\check{A}}_{p+1}^k\cdot\mathbf{\hat{A}}_{p+1}^{k^\top}))
\end{equation}
where $\mathbf{A}_{p+1}^{k}$ is the adjacency matrix of $G_{p+1}^{k}$. 
The subtraction term, tanh, and RELU activation functions may regularize the adjacency matrix so that if the element in $\mathbf{A}_{p+1}^{k}$ is positive, its diagonal counterpart element will be zero.
To strictly force the $G_{p+1}^{k}$ to be unidirectional and acyclic, we adopt the following exponential trace function as constraints inspired by~\cite{zheng2018dags}.
\begin{equation}
    h(\mathbf{A}_{p+1}^{k}) = tr(e^{\mathbf{A}_{p+1}^{k}\circ {\mathbf{A}_{p+1}^{k}}}) - M
\end{equation}
where $\circ$ is the Hadamard product of two matrices and $M$ is the number of nodes (\textit{i.e.}, system entities). 
This function satisfies $h(\mathbf{A}_{p+1}^{k})=0$ if and only if $\mathbf{A}_{p+1}^{k}$ is acyclic.

\noindent\textbf{Optimization.}
To generate a robust casual graph for the new data batch, we jointly optimize all the preceding loss functions. Thus, the final optimization objective is defined as: 
\begin{equation}
\begin{aligned}
& \mathcal{L} = \mathcal{L}_{\hat{G}} + \mathcal{L}_{\check{G}} + \mathcal{L}_{\tilde{p}}  + \mathcal{L}_{\hat{p}} + 
    \mathcal{L}_{\check{p}} 
    \\&
+\lambda_1\cdot(\left \| \mathbf{\hat{A}}_{p+1}^k \right \|_1+\left \| \mathbf{\check{A}}_{p+1}^k \right \|_1)
+\lambda_2\cdot h(\mathbf{A}_{p+1}^{k})
\end{aligned}
\end{equation}
where $\left \| \cdot \right \|_1 $ is the $L1$-norm, which is used to increase the sparsity of $\hat{G}_{p+1}^k$ and $\check{G}_{p+1}^k$to reduce the computational cost; $\lambda_1$ and $\lambda_2$ control the penalized degree of regularization items.

\noindent\textbf{Model Convergence.}
The discovered causal structure and the associated root cause list may gradually converge as the number of new data batches increases.
So we incorporate them as an indicator to automatically terminate the online RCA to avoid unnecessary computing resource waste. 
Assume two consecutive causal graphs are $G_{p+1}^{K-1}$, $G_{p+1}^{K}$, and the associated root cause lists are $\mathbf{l}_{p+1}^{K-1}$, $\mathbf{l}_{p+1}^{K}$.

The node set of the two causal graphs is fixed.
The edge distribution should be comparable when the causal graph converges.
Thus, we define the graph similarity $\varsigma_G$ using the Jensen-Shannon divergence~\cite{fuglede2004jensen} as follows:
\begin{equation}
\varsigma_G = 1 - \text{JS}(P(G_{p+1}^{K-1}) || P(G_{p+1}^{K}))
\end{equation}
where $P(.)$ refers to the edge distribution of the corresponding graph.
$\varsigma_G$ has a value range of $[0\sim 1]$.
The greater the value of $\varsigma_G$ is, the closer the two graphs are.

We use the rank-biased overlap metric~\cite{webber2010similarity} (RBO) to calculate the similarity between two root cause lists in order to fully account for the changing trend of root cause rank.
The ranked list similarity $\varsigma_\mathbf{l}$ is defined as: 
\begin{equation}
    \varsigma_\mathbf{l} = RBO(\mathbf{l}_{p+1}^{K-1},\mathbf{l}_{p+1}^{K})
\end{equation}
$\varsigma_\mathbf{l}$  has a value range of $[0\sim 1]$. The greater the value of $\varsigma_\mathbf{l}$ is, the more similar the two root cause lists are.
We use a hyperparameter $\alpha \in [0\sim 1]$ to integrate $\varsigma_G$ and $\varsigma_\mathbf{l}$, defined as: 
\begin{equation}
    \varsigma = \alpha\cdot\varsigma_G + (1-\alpha)\cdot\varsigma_\mathbf{l}
\end{equation}
The online RCA process may stop when $\varsigma$ is more than a threshold.

\vspace{-0.1cm}
\subsection{Network Propagation based Root Cause Localization}
\label{np_section}
After obtaining the causal graph $G_{p+1}^k$, there are two kinds of nodes: system entities and KPI in the graph. However, the system entities linked to the KPI may not always be the root causes. This is because the malfunctioning effects will spread to neighboring entities starting from the root causes~\cite{cheng2016ranking,dong2017efficient,lin2018collaborative}. We present a random walk-based method for capturing such patterns and more precisely locating root causes.
For simplicity, in this subsection, we directly use $G$ to represent $G_{p+1}^k$.
To trace back the root causes, we first transpose the learned causal graph to get $G^\top$, then adopt a random walk with restarts on the transposed causal graph to estimate the probability score of each entity by starting from the KPI node.

Specifically, we assume that the transition probabilities of a particle on the transposed structure may be represented by $\mathbf{H}$, which has the same shape as the adjacency matrix of $G^\top$. 
Each element in $\mathbf{H}$ indicates the transition probability between any two nodes.
Imagine that from the KPI node, a particle begins to visit the causal structure.
It jumps to any one node with a probability value $\phi \in [0,1]$ or stays at the original position with $1-\phi$.
The higher the value of $\phi$ is, the more possible the jumping behavior happens.
Specifically, if the particle moves from node $i$ to node $j$, the moving probability in $H$ should be updated by: 
\begin{equation}
    \mathbf{H}[i,j] = (1-\phi) \mathbf{A}^\top[i,j] / \sum_{\kappa=1}^M \mathbf{A}^\top [i,\kappa]
\end{equation}
where $\mathbf{A}^\top$ is the adjacency matrix of $G^\top$. 
During the visiting exploration process, we may restart from the KPI node to revisit other entities with the probability $\varphi \in [0,1]$. 
Thus, the visiting probability transition equation of the random walk with restarts can be formulated as: 
\begin{equation}
    \mathbf{q}_{\tau+1} = (1-\varphi)\cdot \mathbf{q}_{\tau} + \varphi \cdot \mathbf{q}_{\xi}
\end{equation}
where $\mathbf{q}_{\tau}$ and $\mathbf{q}_{\tau+1}$ are the visiting probability distributions at the $\tau$ and $\tau+1$ steps, respectively.
$\mathbf{q}_{\xi}$ is the initial visiting probability distribution at the initial step.
When the visiting probability distribution converges, 
the probability scores of the nodes are used as their causal scores to rank them.
The top-$K$ ranked nodes are the most likely root causes of the associated system fault.

\section{Experiments}

\subsection{Experimental Setup}
\subsubsection{Datasets}
We conducted extensive experiments on three real-world datasets:
\noindent\textbf{1) AIOps}: was collected from a real micro-service system that consists of $5$ servers with $234$ pods. From May 2021 to December 2021, the operators collected metric data (\textit{e.g.}, CPU utilization or memory consumption) of all system entities. During this time period, there are 5 system faults. 
\noindent\textbf{2) Swat~\cite{mathur2016swat}:} was collected over an 11-day period from a water treatment testbed that was equipped with $51$ sensors. 
The system had been operating normally for the first 7 days before being attacked in the last 4 days.
There are 16 system faults in the data collection period. 
\noindent\textbf{3) WADI~\cite{ahmed2017wadi}:} was collected from a water treatment testbed over the operation of 16 days.
The testbed consists of $123$ sensors/actuators. 
The system had been operating normally for the first 14 days before being attacked in the last 2 days.
There are $15$ system faults in the collection period.

\subsubsection{Evaluation Metrics}
We compared \model\ with the following four widely-used metrics~\cite{meng2020localizing,liu2021microhecl,guo2018dialog}:
\noindent 1) \textbf{Precision@K (PR@K)}.
It denotes the probability that the top-$K$ predicted root causes are real, defined as:
$
    \text{PR@K} = \frac{1}{\mathbb{|A|}}\sum_{a\in \mathbb{A}}\frac{\sum_{i<K}R_a(i)\in V_a}{min(K,|V_a|)},
$
where $\mathbb{A}$ is the set of system faults; $a$ is one fault in $\mathbb{A}$; $V_a$ is the real root causes of $a$; $R_a$ is the predicted root causes of $a$; and $i$ refers to the $i$-th predicted cause of $R_a$. 
\noindent 2) \textbf{Mean Average Precision@K (MAP@K)}.
It assesses the model performance in the top-$K$ predicted causes from the overall perspective, defined as:
$
    \text{MAP@K} = \frac{1}{K|\mathbb{A}|} \sum_{a\in \mathbb{A}} \sum_{1 \leq j \leq K} \text{PR@j},
$
where a higher value indicates better performance. 
\noindent 3) \textbf{Mean Reciprocal Rank (MRR)}.
This metric measures the ranking capability of models. 
The larger the MRR value is, the further ahead the predicted positions of the root causes are; thus, operators can find the real root causes more easily. 
MRR is defined as:
$
    \text{MRR} = \frac{1}{\mathbb{A}} \sum_{a\in \mathbb{A}} \frac{1}{rank_{R_a}},
$
where $rank_{R_a}$ is the rank number of the first correctly predicted root cause for system fault $a$.
\noindent 4) \textbf{Ranking Percentile (RP).}
Since \model\ is an online root cause localization framework, we expect the ranks of ground-truth root causes to rise as the learning process progresses.
Here, we propose the ranking percentile to evaluate the learning process, defined as:
$
    \text{RP}= (1-\frac{rank_{R_a}}{N})\times 100\% 
    \quad\textit{s.t.}\quad a\in \mathbb{A},
$
where $N$ is the total number of nodes (\textit{i.e.}, system entities).

\begin{table*}[!ht]
\renewcommand\arraystretch{1}
	\centering
	\caption{Overall performance {\it w.r.t.} Swat dataset.} 
	\vspace{-0.3cm}
	\begin{tabular}{c|c|c|c|c|c|c|c|c|c|c}
		\hline
		 & PR@1 & PR@3 & PR@5    & PR@7 & PR@10   & MAP@3      & MAP@5 & MAP@7     & MAP@10 & MRR        \\ \hline
\model & 6.25\% &  \textbf{31.25\%}   & \textbf{55.21\%} & \textbf{64.58\%} & \textbf{92.71\%} & \textbf{15.63\%}  & \textbf{29.79\%} & \textbf{39.73\%} & \textbf{53.96\%} &     31.72\%        \\
NOTEARS & 6.25\% &  7.29\%   & 12.50\% & 39.58\% & 47.92\% & 7.64\%  & 9.58\% & 16.96\% & 25.00\% &     22.36\%        \\
GOLEM & \textbf{18.75\%} &  7.29\%   & 18.75\% & 54.17\% & 62.50\% & 11.81\%  & 13.33\% & 22.02\% & 33.44\% &     30.42\%        \\
Dynotears & \textbf{18.75\%} &  25.00\%   & 29.17\% & 41.67\% & 58.33\% & 23.96\%  & 26.04\% & 29.17\% & 37.08\% &     \textbf{33.99\%}       \\
PC & 12.50\% &  21.88\%   & 36.46\% & 47.92\% & 53.13\% & 19.79\%  & 26.04\% & 31.40\% & 37.40\% &     32.27\%        \\
C-LSTM & 12.50\% &  27.08\%  & 27.08\%  & 39.58\% & 60.42\% & 19.44\% & 22.50\%  & 26.49\% & 34.17\% &    32.86\%        \\
NOTEARS$^*$ & 6.25\% &  29.17\%   & 36.46\% & 55.21\% & 67.71\% & 14.93\%  & 23.54\% & 32.59\% & 42.19\% &     26.30\%        \\
GOLEM$^*$ & 6.25\% &  29.17\%   & 42.71\% & 57.29\% & 68.75\% & 17.01\%  & 26.04\% & 34.97\% & 43.65\% &     28.09\%        \\
\hline
	\end{tabular}
	\vspace{-0.2cm}
	\label{tab:overall_swat}
\end{table*}

\begin{table*}[!ht]
\renewcommand\arraystretch{1}
	\centering
	\caption{Overall performance {\it w.r.t.} WADI dataset.} 
		\vspace{-0.3cm}
	\begin{tabular}{c|c|c|c|c|c|c|c|c|c|c}
		\hline
		 & PR@1 & PR@3 & PR@5    & PR@7 & PR@10   & MAP@3      & MAP@5 & MAP@7     & MAP@10 & MRR        \\ \hline
\model & \textbf{35.71\%} &  \textbf{23.81\%}   & \textbf{60.00\%} & \textbf{70.24\%} & \textbf{83.33\%} & \textbf{28.71\%}  & \textbf{36.05\%} & \textbf{45.82\%} & \textbf{56.00\%} &     \textbf{51.90\%}        \\
NOTEARS & 7.14\% &  23.81\%   & 30.00\% & 35.71\% & 41.67\% & 17.46\%  & 22.55\% & 26.14\% & 30.80\% &     30.12\%        \\
GOLEM & 7.14\% &  10.71\%   & 40.00\% & 51.19\% & 64.29\% & 9.52\%  & 20.14\% & 28.33\% & 38.05\% &     25.89\%        \\
Dynotears & 14.29\% &  29.76\%   & 32.86\% & 42.86\% & 46.43\% & 20.63\%  & 25.38\% & 29.35\% & 33.76\% &     34.28\%        \\
PC & 14.29\% &  21.43\%   & 35.71\% & 45.24\% & 57.14\% & 16.67\%  & 24.29\% & 28.91\% & 35.71\% &     30.74\%        \\
C-LSTM & 12.50\% &  21.43\%   & 46.43\% & 52.38\% & 64.29\% & 17.86\%  & 27.86\% & 34.69\% & 42.62\% &     33.28\% \\
NOTEARS$^*$ & 14.29\% &  20.24\%   & 45.71\% & 66.67\% & 72.62\% & 18.65\%  & 27.48\% & 38.67\% & 48.38\% &     37.74\% \\
GOLEM$^*$ & 21.43\% &  20.24\%   & 60.00\% & 64.29\% & 73.81\% & 19.84\%  & 30.33\% & 39.86\% & 48.98\% &     40.24\% \\
\hline
	\end{tabular}
	\vspace{-0.2cm}
	\label{tab:overall_wadi}
\end{table*}

\begin{table*}[!ht]
\renewcommand\arraystretch{1}
	\centering
	\caption{Overall performance {\it w.r.t.} AIOps dataset.} 
		\vspace{-0.3cm}
	\begin{tabular}{c|c|c|c|c|c|c|c|c|c|c}
		\hline
		 & PR@1 & PR@3 & PR@5    & PR@7 & PR@10   & MAP@3      & MAP@5 & MAP@7     & MAP@10 & MRR        \\ \hline
\model & \textbf{80.00\%} &  \textbf{100.0\%}   & \textbf{100.0\%} & \textbf{100.0\%} & \textbf{100.0\%} & \textbf{93.33\%}  & \textbf{96.00\%} & \textbf{97.14\%} & \textbf{98.00\%} &     \textbf{90.00\%}        \\
NOTEARS & 0.00\% &  40.00\%   & 80.00\% & 40.00\% & 60.00\% & 20.00\%  & 28.00\% & 37.14\% & 44.00\% &     20.46\%        \\
GOLEM & 20.00\% &  60.00\%   & 60.00\% & 60.00\% & 60.00\% & 40.00\%  & 40.00\% & 51.43\% & 54.00\% &     37.74\%        \\
Dynotears & 40.00\% &  60.00\%   & 60.00\% & 60.00\% & 60.00\% & 53.33\%  & 56.00\% & 57.14\% & 58.00\% &     50.77\%        \\
PC & 20.00\% &  20.00\%   & 20.00\% & 40.00\% & 60.00\% & 20.00\%  & 20.00\% & 22.86\% & 30.00\% &    25.36\%        \\
C-LSTM & 0.00\% &  40.00\%   & 60.00\% & 60.00\% & 60.00\% & 26.67\%  & 36.00\% & 42.86\% & 48.00\% &    24.73\%        \\
NOTEARS$^*$ & 40.00\% &  80.00\%   & 80.00\% & 80.00\% & 80.00\% & 66.67\%  & 72.00\% & 74.29\% & 76.00\% &    60.00\%        \\
GOLEM$^*$ & 60.00\% &  80.00\%   & 80.00\% & 80.00\% & 80.00\% & 73.33\%  & 76.00\% & 77.14\% & 78.00\% &    70.00\%        \\
\hline
	\end{tabular}
	\label{tab:overall_aiops}
\end{table*}

\begin{figure*}[htbp]
\centering
\subfigure[AIOps 0901]{
\includegraphics[width=4.3cm]{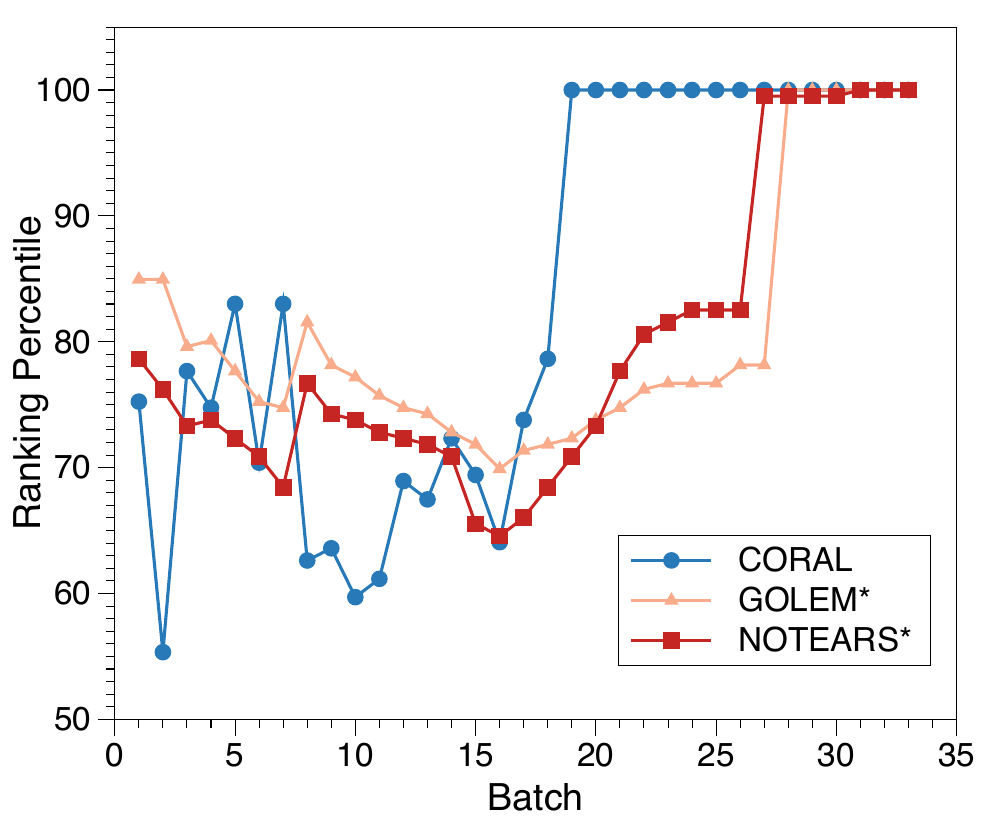}
}
\hspace{-3mm}
\subfigure[AIOps 0524]{ 
\includegraphics[width=4.3cm]{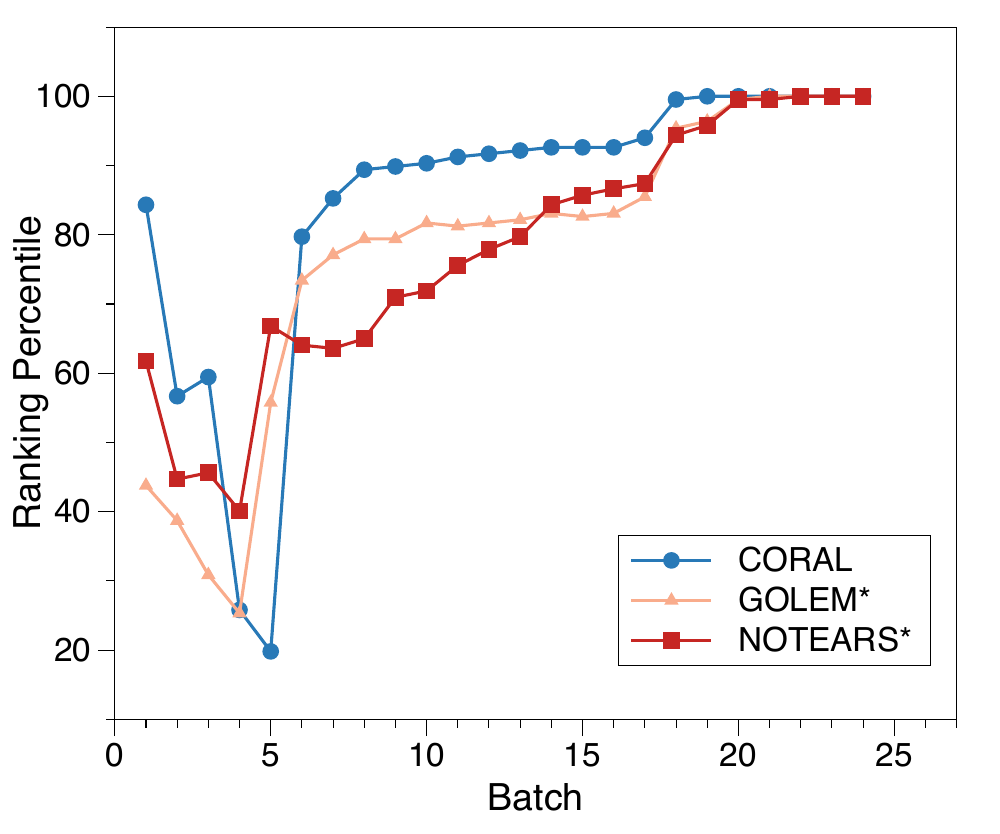}
}
\hspace{-3mm}
\subfigure[Swat Fault 8]{ 
\includegraphics[width=4.3cm]{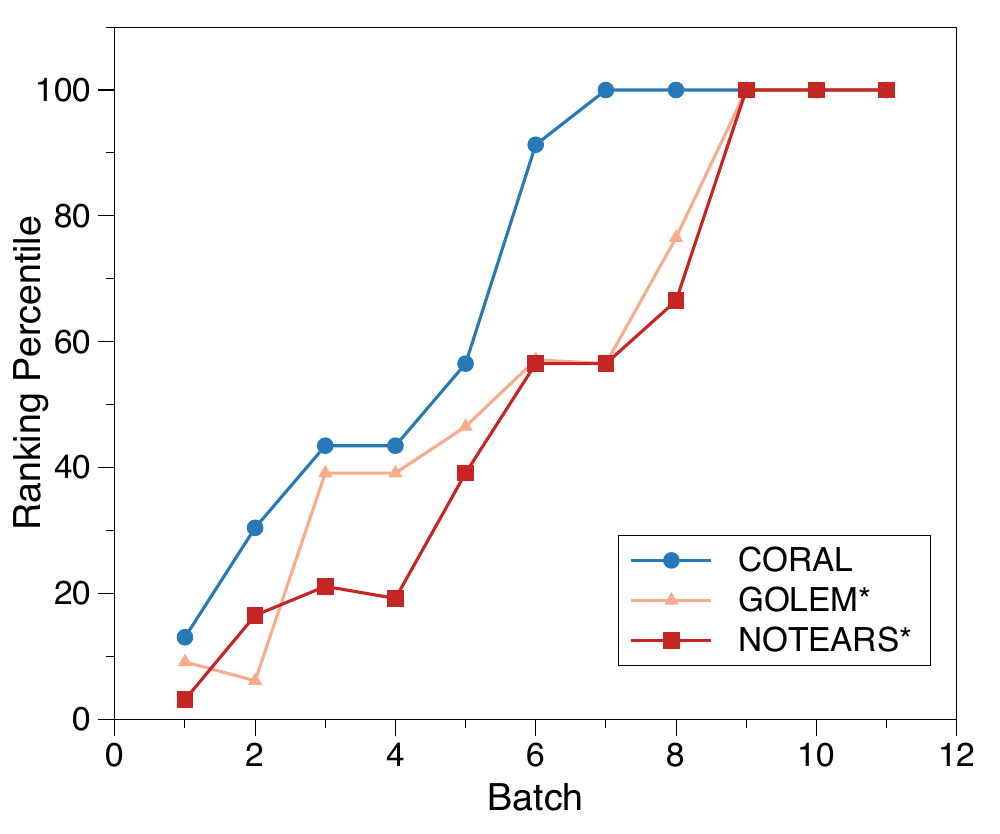}
}
\hspace{-3mm}
\subfigure[WADI Fault 6]{ 
\includegraphics[width=4.3cm]{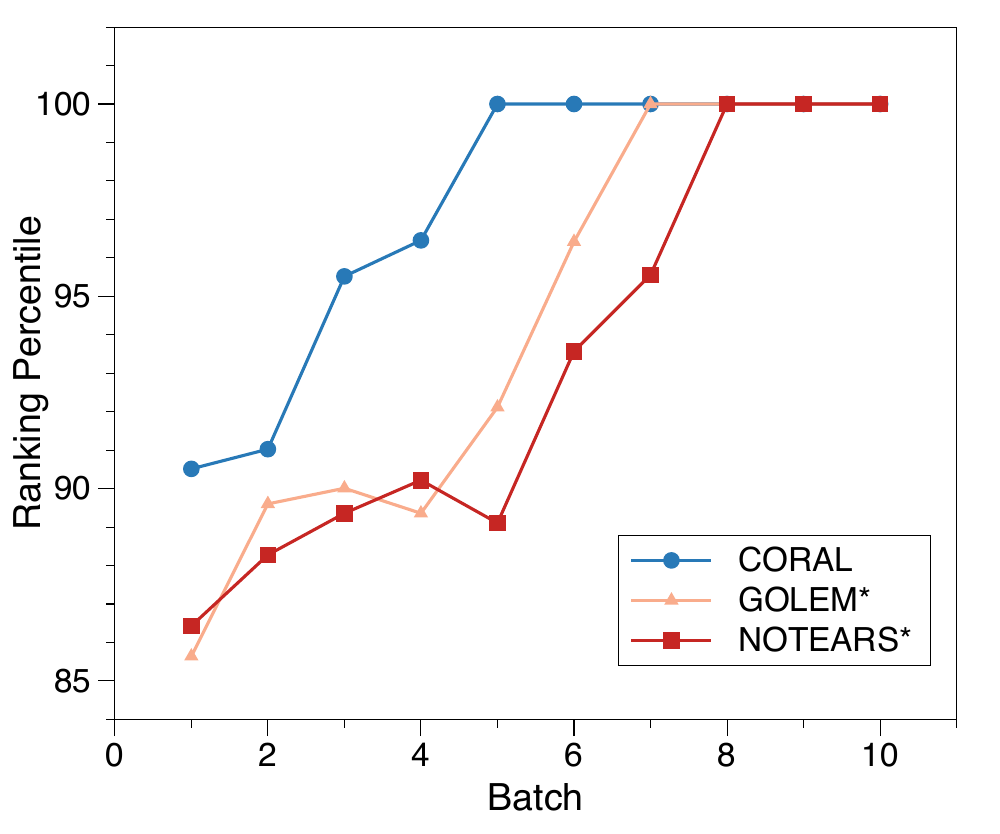}
}
\vspace{-0.3cm}
\caption{Comparison of online RCA models on different batches in terms of ranking percentile.}
\label{search}
\end{figure*}

\subsubsection{Baselines}
We compared \model\ with the following five causal discovery models:
\nop{\textbf{(1) NOTEARS~\cite{zheng2018dags}}  is fundamentally distinct from prior heuristics-based search-based approaches, which formulate the structure learning problem as a purely continuous constrained optimization problem over real matrices to increase learning efficiency.
\textbf{(2) GOLEM~\cite{ng2020role}} proposes a likelihood-based score function to relax the hard DAG constraint in NOTEARS for easily discovering the causal structure among thousands of nodes.
\textbf{(3) PC\cite{spirtes2000causation}} belongs to the constraint-based technical category, which  first identifies the skeleton of the causal graph using  the conditional independence test and  then determines the causal direction using the Markov equivalence class and acyclicity constraints.
\textbf{(4) C-LSTM~\cite{tank2021neural}}
utilizes LSTM neural networks combined with sparsity-inducing penalties on the weights to extract non-linear Granger causality from time series data.
\textbf{(5) Dynotears~\cite{pamfil2020dynotears}}
belongs to the score-based category, which discovers dynamic Bayesian networks by minimizing a penalized loss subject to an acyclicity constraint to capture contemporaneous and time-lagged causal relations simultaneously.
}
1) \textbf{NOTEARS}~\cite{zheng2018dags} 
formulates the structure learning problem as a purely continuous constrained optimization problem over real matrices to increase learning efficiency.
 2) \textbf{GOLEM}~\cite{ng2020role} employs a likelihood-based score function to relax the hard DAG constraint in NOTEARS.
3) \textbf{Dynotears}~\cite{pamfil2020dynotears} is a score-based method that uses the structural vector autoregression model to construct dynamic Bayesian networks.
4) \textbf{PC}~\cite{spirtes2000causation} is a classic constraint-based method. It first identifies the skeleton of the causal graph with the independence test, then generates the orientation direction using the v-structure and acyclicity constraints.
5) \textbf{C-LSTM}~\cite{tank2021neural} captures the nonlinear Granger causality that existed in multivariate time series by using LSTM neural networks.

All these models can only learn the causal structure from time series data offline. 
Thus, we first collect monitoring data from the beginning until system failures occur as historical records.
Then, based on the collected records, we apply the causal discovery models to learn causal graphs and leverage network propagation on such graphs to identify the top $K$ nodes as the root causes.
Besides, since \model\ is an online RCA framework, we extend NOTEARS and GOLEM to the online learning setting, denoted by \textbf{NOTEARS$^*$} and \textbf{GOLEM$^*$}, respectively, for a fair comparison\footnote{\tiny{Other baselines are not extended to the online setting as they are time-intensive when there are multiple data batches.}}.
For the online setting, we collect monitoring data before the first trigger point as training or historical data to construct the initial causal graph, which describes the normal system state. Once a trigger point is detected, \nop{we first build an initial causal graph that depicts the normal system state using historical data.
Then, the trigger point detector is deployed to detect system transition points.
If detected,}we update the causal graph iteratively for each new batch of data. \model\ can inherit the causations from the previous data batch.
NOTEARS$^*$ and GOLEM$^*$ have to learn from scratch for each new data batch.
All causal discovery baseline models are implemented using the gcastle library\footnote{\tiny{https://github.com/huawei-noah/trustworthyAI/tree/master/gcastle.}}.

\subsubsection{Experimental Settings}
All experiments are conducted on a server running Ubuntu 18.04.5 with Intel(R) Xeon(R) Silver 4110 CPU @ 2.10GHz, 4-way GeForce RTX 2080 Ti GPUs, and 192 GB memory.
In addition, all methods were implemented using Python 3.8.12 and PyTorch 1.7.1.

\begin{figure*}[htbp]
\centering
\subfigure[AIOps 0901]{
\includegraphics[width=4.3cm]{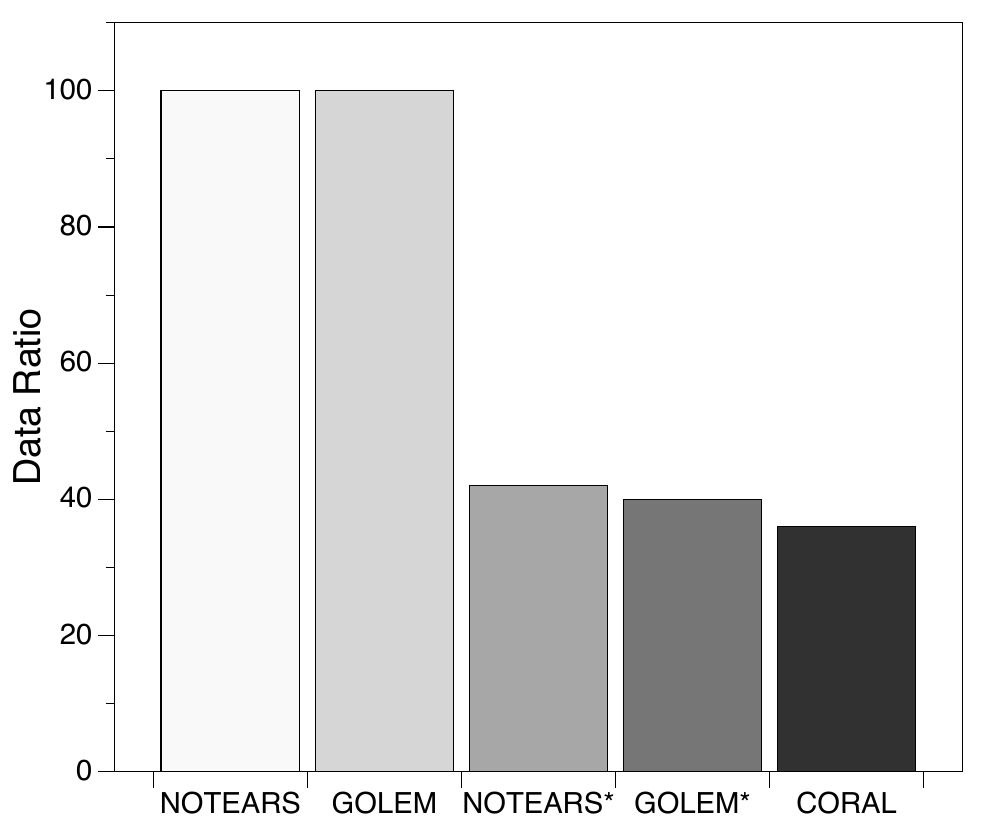}
}
\hspace{-3mm}
\subfigure[AIOps 0524]{ 
\includegraphics[width=4.3cm]{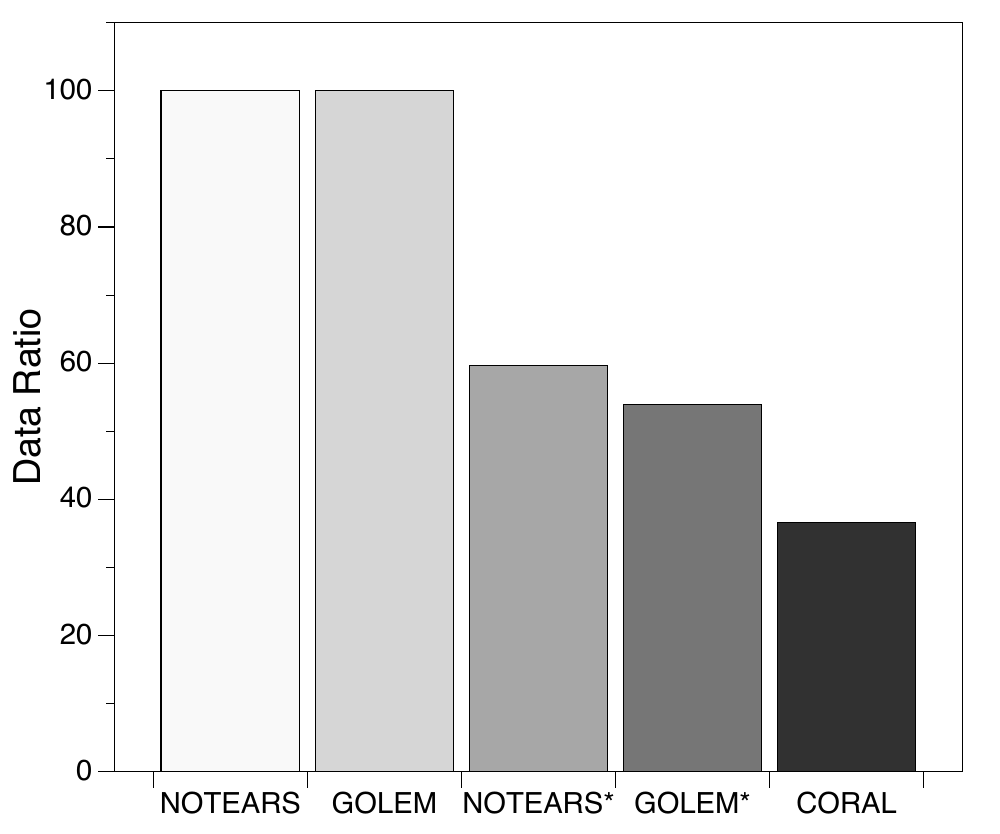}
}
\hspace{-3mm}
\subfigure[Swat Fault 8]{ 
\includegraphics[width=4.3cm]{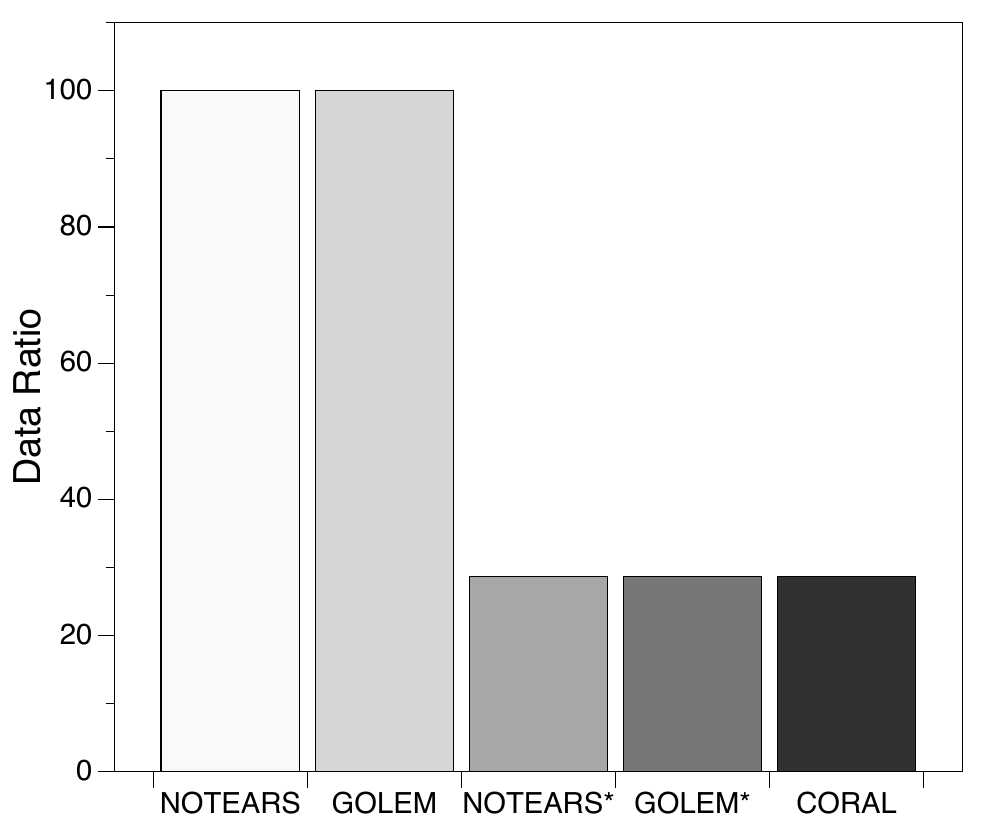}
}
\hspace{-3mm}
\subfigure[WADI Fault 6]{ 
\includegraphics[width=4.3cm]{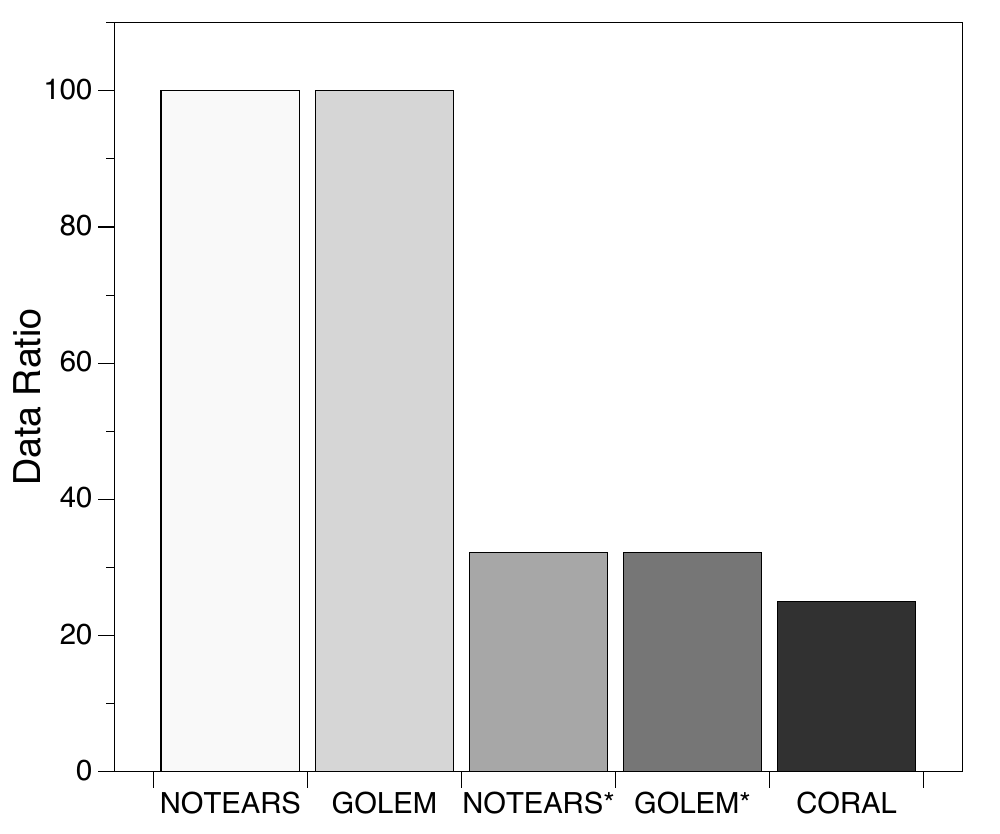}
}
\vspace{-0.4cm}
\caption{Comparison of required data volume for different root cause analysis models.}
\label{data_ratio}
\end{figure*}

\begin{figure*}[htbp]
\vspace{-0.3cm}
\centering
\subfigure[SWAT MAP@10]{
\includegraphics[width=4.3cm]{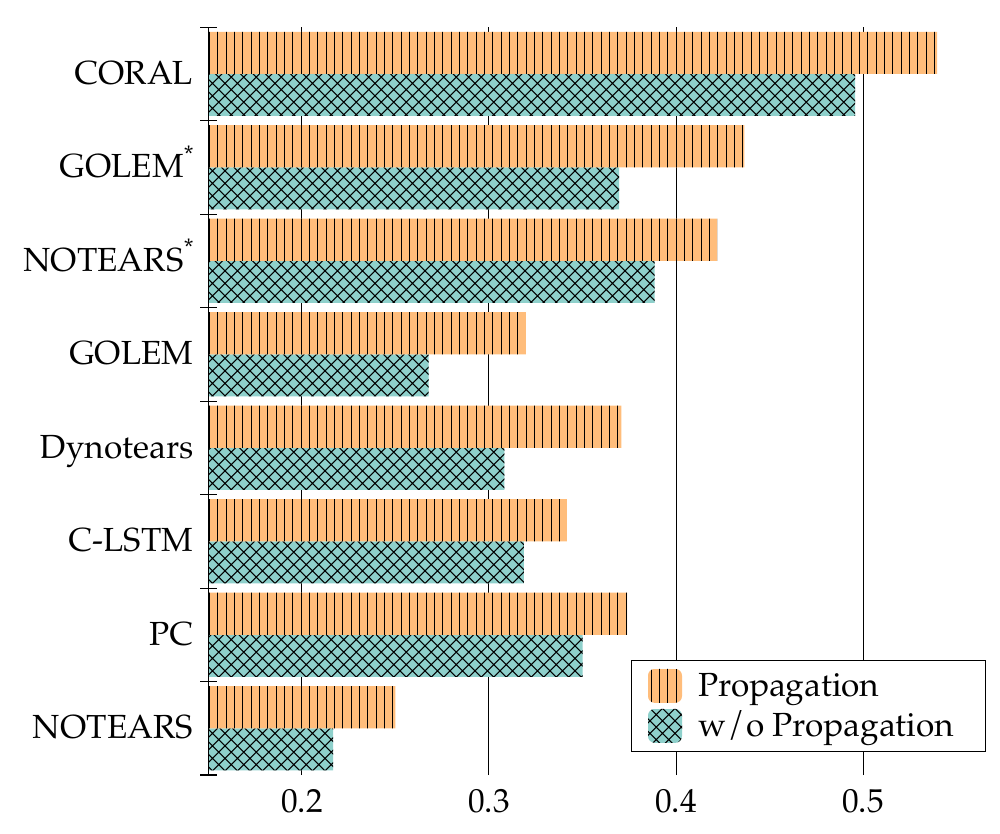}
}
\hspace{-3mm}
\subfigure[SWAT MRR]{ 
\includegraphics[width=4.3cm]{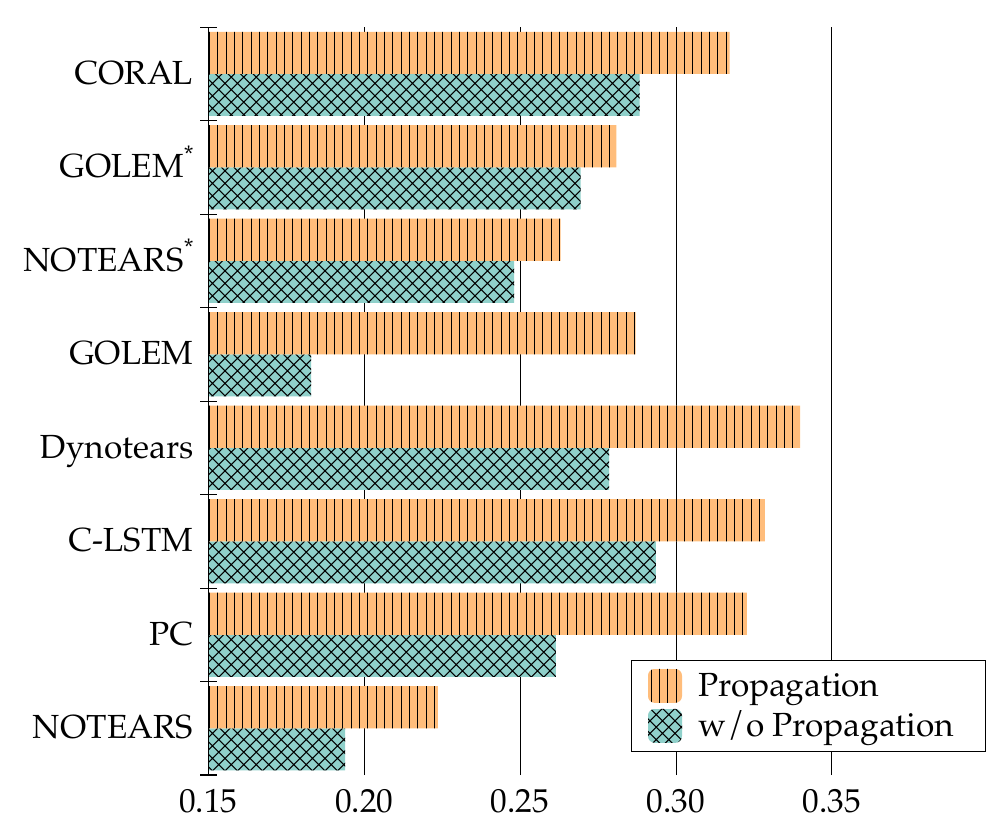}
}
\hspace{-3mm}
\subfigure[WADI MAP@10]{ 
\includegraphics[width=4.3cm]{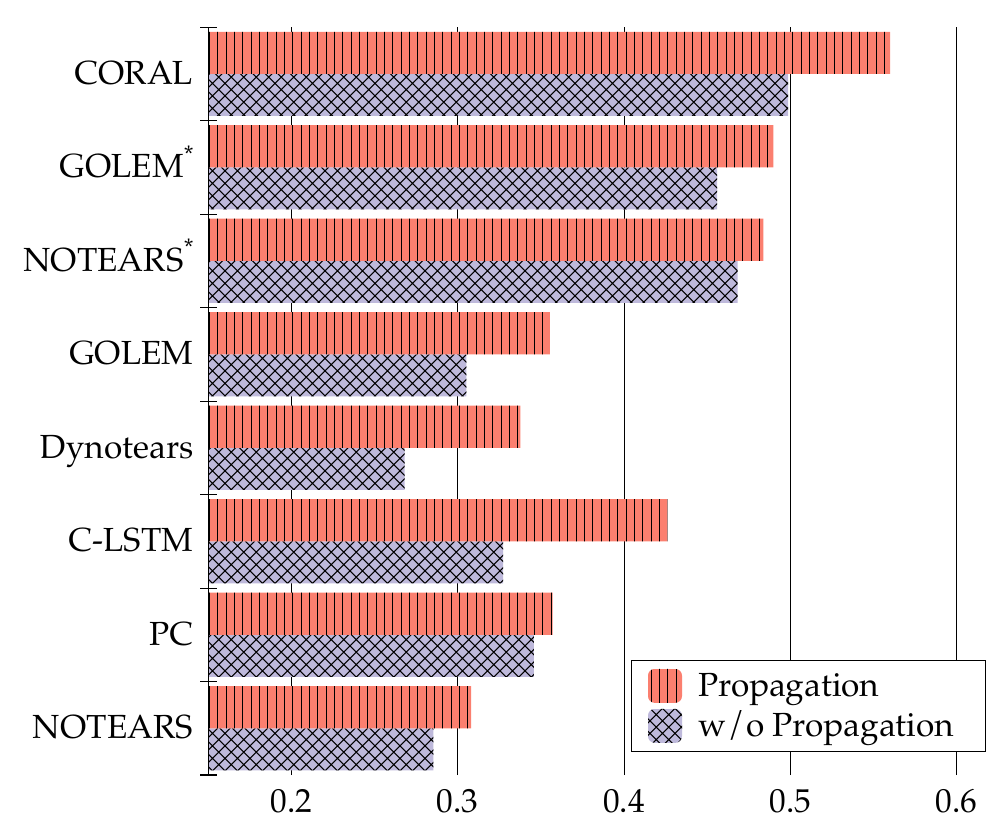}
}
\hspace{-3mm}
\subfigure[WADI MRR]{ 
\includegraphics[width=4.3cm]{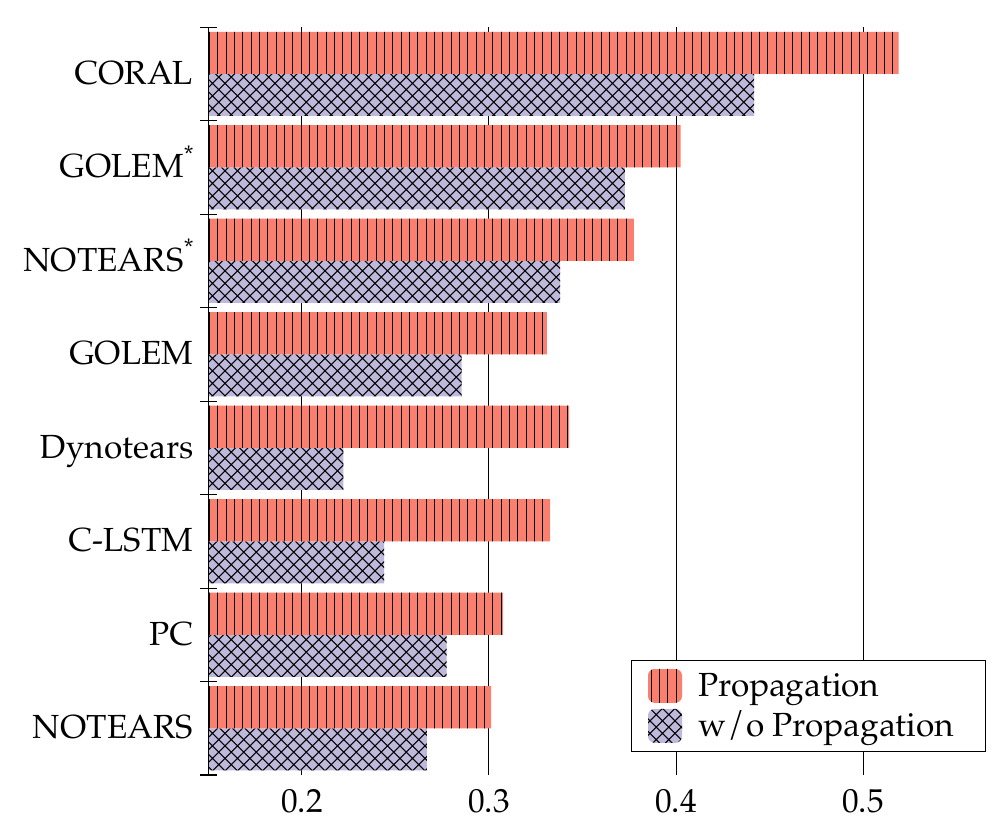}
}
\vspace{-0.4cm}
\caption{Study of the impact of network propagation for RCA.}
\label{np_figure}
\end{figure*}

\vspace{-0.1cm}
\subsection{Performance Evaluation}

\subsubsection{Overall Performance}
Table~\ref{tab:overall_swat}, Table~\ref{tab:overall_wadi}, and Table~\ref{tab:overall_aiops} show the overall performance of all models.
The greater the value of evaluation measures, the superior the model's performance.
There are two key observations: 
First, the online methods (\textit{e.g.}, NOTEARS$^*$, GOLEM$^*$, \model) outperform the offline approaches (\textit{e.g.}, Dynotears, PC, and C-LSTM).
The underlying driver is that online methods can rapidly capture the changing patterns in the monitoring metric data. So, they can learn an accurate and noise-free causal structure for RCA. Second, compared to online methods, \model\ still significantly outperforms NOTEARS$^*$ and GOLEM$^*$.
The underlying reason is that the disentangled causal graph learning module independently learns state-invariant and state-dependent causal relations for incrementally updating the causal graph.
Such a learning manner can learn more robust and accurate causal relations, enhancing root localization performance. 
Thus, the experimental results demonstrate the superiority and effectiveness of \model\ in root cause localization over other baselines.

\subsubsection{Learning Procedure Analysis.}
Fig.~\ref{search} illustrates how the performance of online RCA frameworks varies as the number of data batches grows. 
As more data comes in, we can see that the ranking percentiles of root causes identified by all online frameworks can converge.
This shows how well the online RCA works, since it can gradually pick up on the changing patterns in monitoring metric data.
Another interesting observation is that \model\ can identify the root causes earlier than NOTEARS$^*$, GOLEM$^*$.
For instance, in the AIOps 0901 case, \model\ can identify the root causes nine batches earlier than the other two models.
A possible reason is that \model\ updates the new causal graph by disentangling the state-invariant and state-dependent information instead of learning from scratch.
It may generate more robust and effective causal structures.
This experiment shows the validity of RCA's online learning setup and the utility of disentangled causal graph learning.

\subsubsection{Required Data Volume Analysis.}
Fig.~\ref{data_ratio} shows the comparison results of required data volumes for different RCA models.
We find that compared with offline models, the online RCA approaches can significantly reduce the required data volume.  
In AIOps 0524, for instance, NOTEARS$^*$ reduces data usage by at least 40\% compared to the offline models.
A possible explanation is that the online RCA setting may capture the system state dynamics in order to acquire a good causal structure without being affected by an excessive amount of redundant data.
Moreover, we observe that \model\ uses fewer data but achieves better results compared with other baselines.
A potential reason is that disentangled causal graph learning may inherit causal information from the previous state, resulting in reduced data consumption while  preserving a robust causal structure and good RCA performance. Thus, this experiment shows that \model\ can reduce the computational costs of RCA.

\subsubsection{Influence of Network Propagation}
Here, we apply our network propagation mechanism (see Section~\ref{np_section}) to the causal structures learned by each causal discovery model to investigate its impact on performance.
For the control group, we eliminate the propagation effect by directly selecting the system entities connected to the system KPI as root causes. Fig.~\ref{np_figure} shows the comparison results of all models on SWAT and WADI data in terms of MAP@10 and MRR.
We can find that network propagation significantly enhances the RCA performance in all cases.
A possible reason for this observation is that in most cases, the malfunctioning effects of root causes may propagate among system entities over time. Network propagation can simulate such patterns in the learned causal structure, resulting in a better RCA performance.
Thus, this experiment demonstrates that the network propagation module is critical to the performance of an RCA model.

\subsubsection{Time Cost Analysis.}
\begin{figure}[!t]
\centering
\subfigure[ AIOps 0524]{
\includegraphics[width=4.2cm]{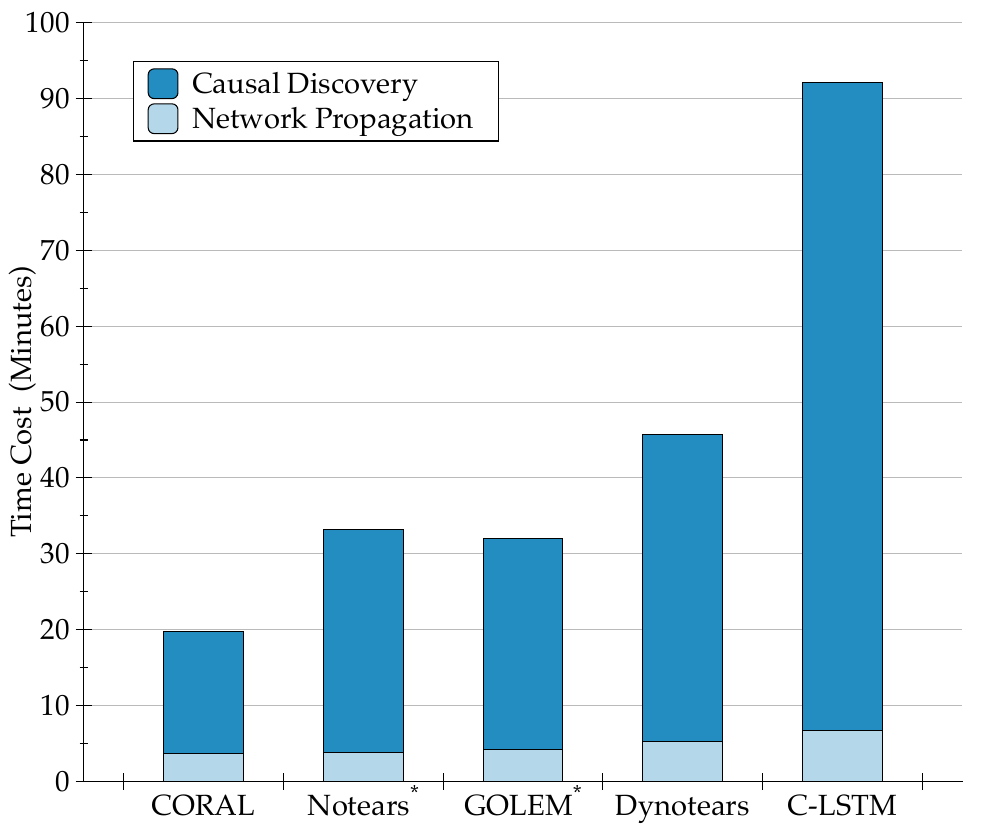}}
\hspace{-3.5mm}
\subfigure[AIOps 0901]{
\includegraphics[width=4.2cm]{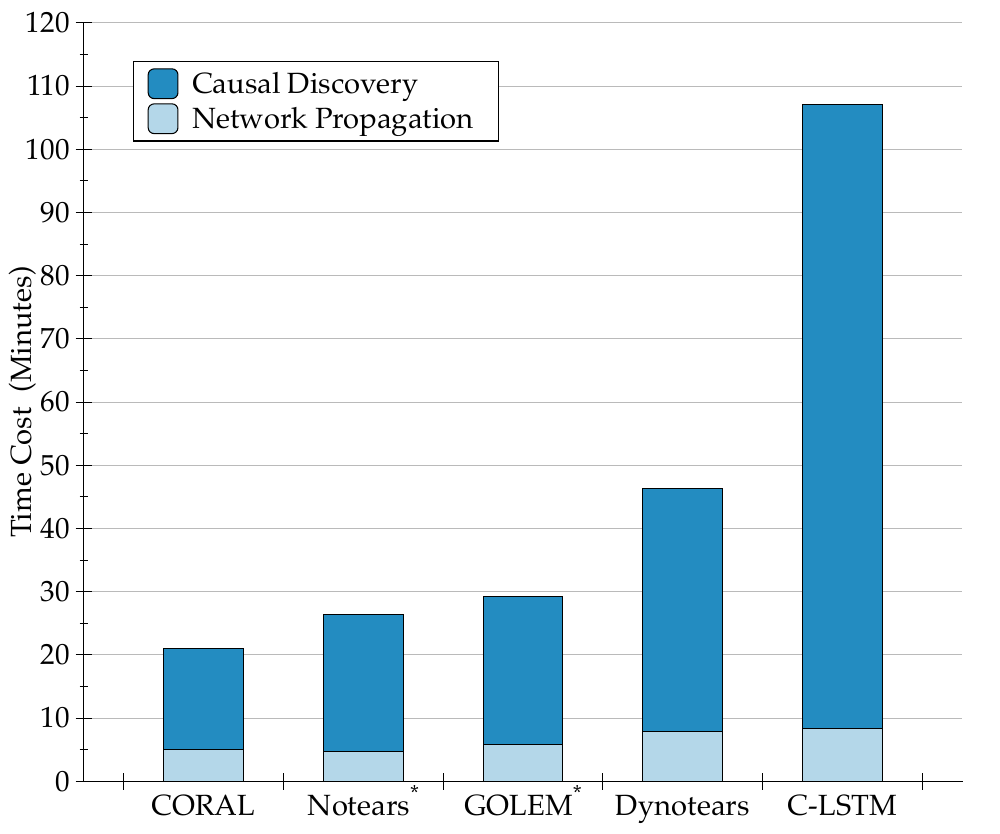}}
\vspace{-0.05cm}
\caption{Comparison of the time costs associated with causal discovery and network propagation of distinct models.}
\label{time_cost}
\end{figure}

Fig.~\ref{time_cost} shows the time cost comparison of causal structure learning and network propagation of distinct models using two fault cases of AIOps, denoted by 0524 and 0901, respectively.
We can observe that offline methods require more time than online methods for network propagation. A potential reason is that offline methods utilize a large amount of historical data for learning causal relations, resulting in a more dense causal network with noisy causal relations than online methods. Thus, it requires more time for the network propagation module of offline methods to converge. Another interesting observation is that \model\ requires the least amount of time in causal structure learning compared with other baselines. The underlying driver is that the incremental causal graph learning module in \model\ maintains the state-invariant causal relations and updates the state-dependent ones, hence accelerating the causal structure learning. 


\begin{figure}[!t]
\vspace{-1.pt}
\centering
\subfigure[ AIOps 0524 (Metric data + KPI)]{
\includegraphics[width=4.2cm]{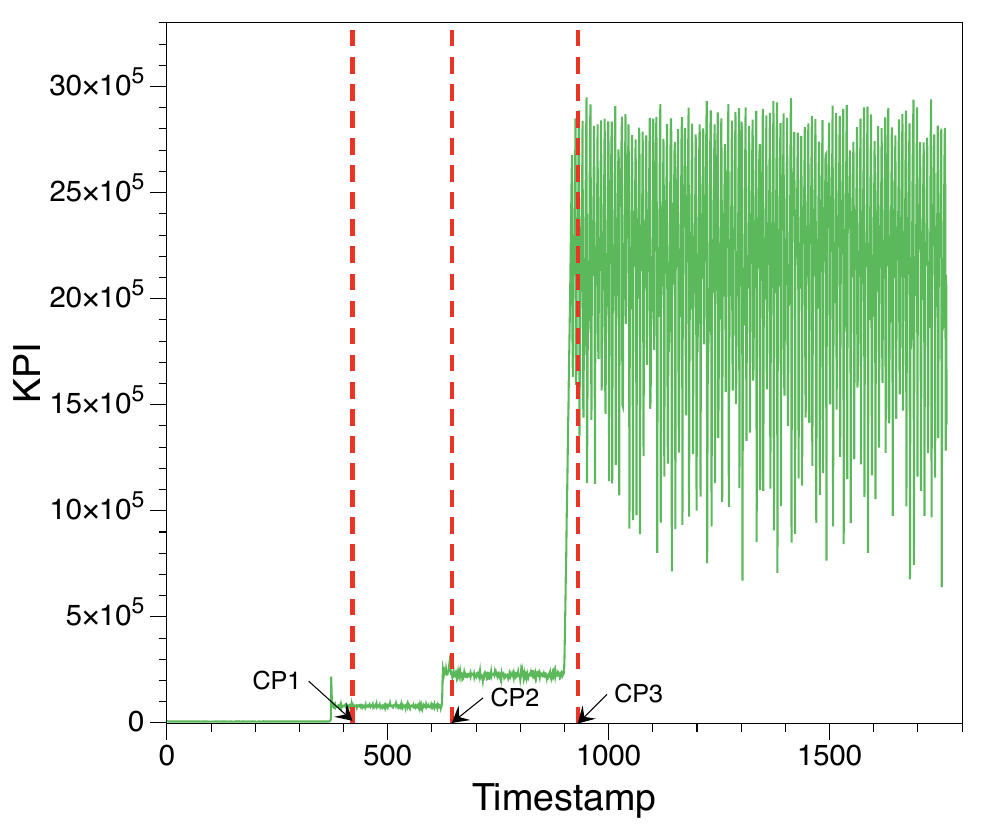}}
\hspace{-3.5mm}
\subfigure[AIOps 0524 (KPI) ]{
\includegraphics[width=4.2cm]{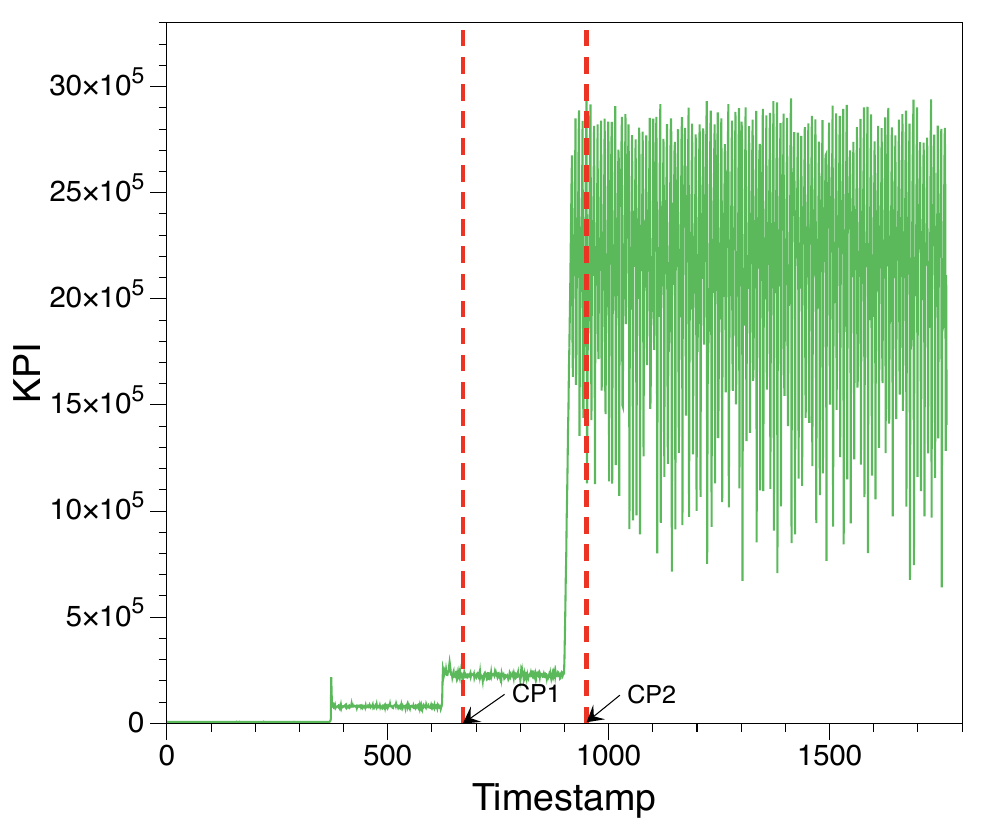}}
\vspace{-0.4cm}
\caption{Comparison of trigger point detection with and w/o metric data. Red dashed lines indicate the trigger points.}
\label{cpd_check}
\end{figure}

\subsubsection{Trigger Point Detection Analysis.}
\nop{We create a trigger point detection module to automatically identify the system state shifts to initiate online RCA  (see Section~\ref{tpd_sec}).}
In our proposed trigger point detection module (see Section~\ref{tpd_sec}), we use both system metric data and KPI to identify the transition points between system states. To evaluate its effect on performance, we use only the system KPI data for determining the trigger points in the control group.
Fig.~\ref{cpd_check} shows the comparison results on AIOps 0524 data by visualizing its system KPI and associated trigger points (highlighted by red dashed lines).
We find that by integrating metric data into KPI, we can detect system transition points more quickly and precisely.
A potential reason is that metric data may contain some early failure symptoms or precursor patterns, which can not be captured in the system key performance indicator.  


\subsubsection{A Case Study.}
To further illustrate the learned state-invariant and state-dependent causal graphs, we conduct a case study using the system failure of AIOps on September 1, 2021.
System operators first deployed a microservice system on three servers: \textit{compute-2}, \textit{infra-1}, and \textit{control-plane-1}. Then, they sent requests periodically to the pod \textit{sdn-c7kqg} to observe the system's latency.
Finally, the operators attacked the \textit{catalogue-xfjp} by causing its CPU load extremely high.
Such an attack event may affect other pods deployed on different servers and lead to a system fault.
During this procedure, \model\ is employed for online root cause localization. 

\begin{figure}[!t]
    \centering
    \includegraphics[width=0.85\linewidth]{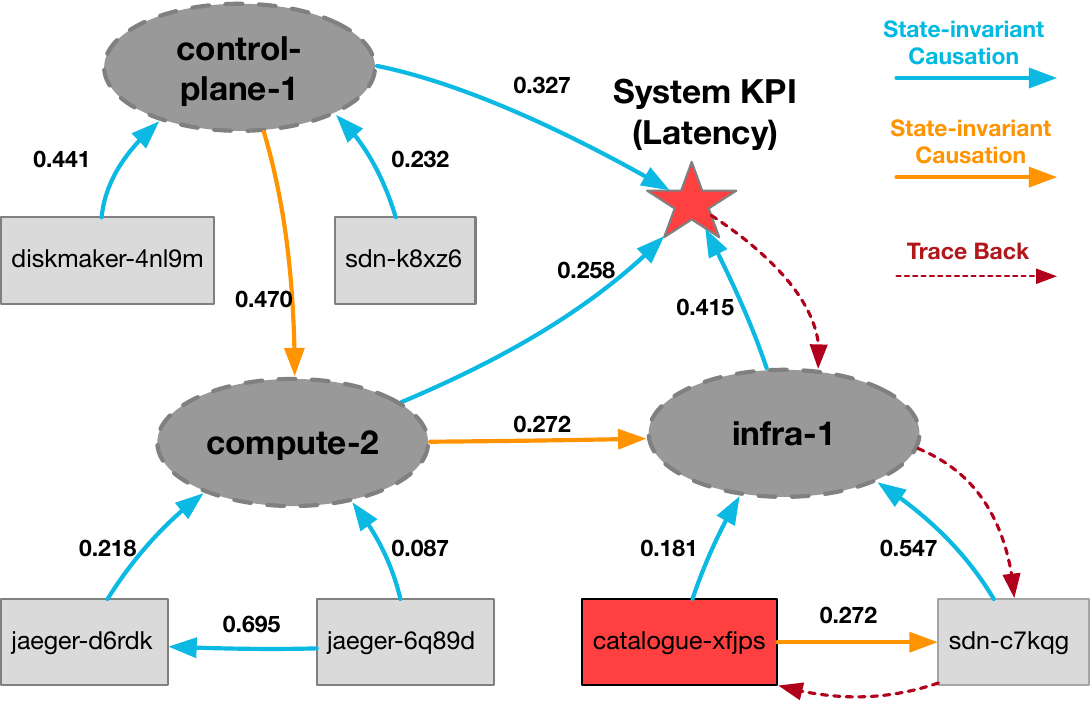}
    \captionsetup{justification=justified, singlelinecheck=off}
    \caption{ State-invariant and state-dependent causal relations learned from AIOps dataset.
    Applications/pods are denoted by solid rectangle nodes, with the red one representing the root cause. 
    Servers are denoted by dashed circles.
    The number on each causal relation edge indicates the causal score between two connected nodes. The red dashed line reflects the backtracing process of the root cause.}
    \label{fig:causal_graph}
\end{figure}

Fig.~\ref{fig:causal_graph} shows the causal graph generated by \model, when the ground-truth root cause is ranked first. The blue and orange arrow lines represent state-invariant and state-dependent causation, respectively.
From Fig.~\ref{fig:causal_graph}, we can first find that \textit{infra-1} server is the most possible one that increases the system latency. Then, using the causal score to trace back from this node, the root cause node \textit{catalogue-xfjp} can be identified.
This observation demonstrates that the \model\ can accurately learn state-invariant and state-dependent causal relationships and provide a mechanism for traceable and explainable root cause analysis.

\section{Related Work}
\noindent\textbf{Root Cause Localization} focuses on identifying the root causes of system failures/faults based on symptom observations~\cite{sole2017survey}.
Many offline RCA approaches~\cite{fourlas2021survey,deng2021graph,tang2019graph,gui2020aptrace,soldani2022anomaly} have been proposed to improve system robustness in various domains.
In energy domains, Capozzoli {\it{et al.}} utilized statistical techniques and DNNs to determine the cause of abnormal energy consumption~\cite{capozzoli2015fault}.
In web development domains, Brandon {\it et al.} proposed a graph representation framework to identify root causes in microservice systems by comparing anomalous circumstances and graphs~\cite{brandon2020graph}. 
Different from existing offline studies, \model\ is an online RCA approach. 

\noindent\textbf{Causal Discovery in Time Series}
aims to discover causal structures using observational time series data~\cite{assaad2022survey}. 
Existing approaches are four-fold: (i) Granger causality approaches~\cite{nauta2019causal,tank2021neural}, in which causality is assessed according to whether one time series is useful for predicting another;
(ii) Constraint-based approaches~\cite{runge2020discovering,sun2015causal}, in which  causal skeleton is first identified using conditional independence test, and then the causal directions are determined based on the Markov equivalence relations;
(iii) Noise-based approaches~\cite{hyvarinen2010estimation,peters2013causal}, in which causal relations among distinct variables are depicted by the combinations between these variables and associated noises; 
(iv) Score-based approaches~\cite{pamfil2020dynotears,bellot2021neural,wang2023hierarchical}, in which the validity of a candidate causal graph is evaluated using score functions, and the graph with the highest score is the final output.
Existing studies assume that the causal graph underlying the time series is constant and stable.
However, real-world causal relationships between different variables are dynamic and vary over time.
Our study proposes a novel incremental causal discovery model based on disentangled graph learning.

\noindent\textbf{Change Point Detection} is to identify the state transitions in time-series data. It can be categorized into offline detection and online detection~\cite{aminikhanghahi2017survey}.
Offline change point detection takes the entire historical time series as input and outputs all possible change points at once.
It has many applications including speech processing~\cite{seichepine2014piecewise}, financial analysis~\cite{frick2014multiscale}, bio-informatics~\cite{liu2018change}.
Online change point detection examines new data once available in order to detect new change points quickly. It can detect change points in real-time or near real-time when  time series are monitored. The method is frequently used to detect the occurrence of major incidents/events (\textit{e.g.}, system faults) in monitoring systems~\cite{8477020,bakdi2021real}.
To rapidly identify state transitions, we employ an online change point detection method using multivariate singular spectrum analysis~\cite{alanqary2021change}.


\noindent\textbf{Disentangled Representation Learning} aims to identify and disentangle the underlying explanatory factors hidden in the observational data~\cite{xin2022}. Disentangled representation Learning has been widely used in a variety of domains, including computer vision~\cite{hsieh2018learning}, time series analysis~\cite{li2022towards,chen2022interpretable}, graph learning~\cite{wang2020disentangled}, and natural language processing~\cite{he2017unsupervised}. In particular, for graph learning, 
Guo \textit{et al.}~\cite{guo2020interpretable} proposed an unsupervised disentangled approach to disentangle node and edge features from attributed graphs. 
To learn graph-level disentangled representations, 
Li \textit{et al.}~\cite{li2021disentangled} presented a method for learning disentangled graph representations with self-supervision. 
Different from the previous works, we identify and disentangle the state-invariant and state-dependent factors 
for incremental causal graph learning. 
\section{Conclusion}
In this paper, we studied a novel and challenging problem of incremental causal structure learning for online root cause analysis (RCA). To tackle this problem, we first designed an online trigger point detection module that can detect system state changes and trigger early RCA. To decouple system state-invariant and system state-dependent factors, we proposed a novel incremental disentangled causal graph learning approach to incrementally update the causal structure for accurate root cause localization. Based on the updated causal graph, we designed a random walk with restarts to model the network propagation of system faults to accurately identify the root cause.
We conducted comprehensive experiments on three real-world datasets to evaluate the proposed framework. The experimental results validate its effectiveness and the importance of incremental causal structure learning and online trigger point detection.
An interesting direction for further exploration would be incorporating other sources of data, such as system logs, with the time series data for online RCA in complex systems. 

\newpage
\bibliographystyle{ACM-Reference-Format}
\bibliography{acmart1}

\end{document}